\documentclass{article}

\usepackage[margin=1in]{geometry}

\usepackage{microtype}
\usepackage{graphicx}
\usepackage{subfigure}
\usepackage{booktabs} 

\usepackage{hyperref}

\usepackage{amsmath,amsthm,amsfonts,amssymb}
\usepackage{mathtools}
\usepackage{xcolor}
\usepackage{natbib}
\usepackage{algorithm} 
\usepackage{algpseudocode} 
\usepackage{url}
\usepackage{multirow}
\usepackage{authblk}

\usepackage[T1]{fontenc}
\usepackage[utf8]{inputenc}
\usepackage{bbm}

\def\*#1{\boldsymbol{#1}}

\theoremstyle{plain}
\newtheorem*{rep@theorem}{\rep@title}
\newcommand{\newreptheorem}[2]{%
\newenvironment{rep#1}[1]{%
 \def\rep@title{#2 \ref{##1}}%
 \begin{rep@theorem}}%
 {\end{rep@theorem}}}

\newtheorem{theorem}{Theorem}[section]
\newreptheorem{theorem}{Theorem}
\newtheorem{lemma}{Lemma}[section]
\newreptheorem{lemma}{Lemma}
\newtheorem{corollary}{Corollary}[section]
\newreptheorem{corollary}{Corollary}
\newtheorem{assumption}{Assumption}[section]
\newreptheorem{assumption}{Assumption}
\newtheorem{definition}{Definition}[section]
\newreptheorem{definition}{Definition}

\newcommand{\myPr}{\mathrm{Pr}}

\DeclareMathOperator*{\argmax}{arg\,max}

\newcommand{\lw}[1]{\smash{\lower2.ex\hbox{#1}}}

\newcommand{\RR}{\mathbb{R}}

\newcommand{\EE}{\mathbb{E}}

\newcommand{\cD}{{\cal D}}

\newcommand{\cG}{{\cal G}}

\newcommand{\cN}{{\cal N}}

\newcommand{\cP}{{\cal P}}

\newcommand{\cX}{{\cal X}}

\mathtoolsset{showonlyrefs}

\title{Randomized Gaussian Process Upper Confidence Bound \\ with Tighter Bayesian Regret Bounds}

\author[1,2]{Shion Takeno}
\author[1]{Yu Inatsu}
\author[1]{Masayuki Karasuyama}

\affil[1]{Nagoya Institute of Technology}
\affil[2]{RIKEN AIP}

\affil[ ]{\texttt{shion.takeno@riken.jp, {\rm \{}inatsu.yu, karasuyama{\rm\}}@nitech.ac.jp}}
\date{}

\begin{document}
\maketitle

\begin{abstract}
  Gaussian process upper confidence bound (GP-UCB) is a theoretically promising approach for black-box optimization; however, the confidence parameter $\beta$ is considerably large in the theorem and chosen heuristically in practice.
  Then, randomized GP-UCB (RGP-UCB) uses a randomized confidence parameter, which follows the Gamma distribution, to mitigate the impact of manually specifying $\beta$.
  This study first generalizes the regret analysis of RGP-UCB to a wider class of distributions, including the Gamma distribution.
  Furthermore, we propose improved RGP-UCB (IRGP-UCB) based on a two-parameter exponential distribution, which achieves tighter Bayesian regret bounds.
  IRGP-UCB does not require an increase in the confidence parameter in terms of the number of iterations, which avoids over-exploration in the later iterations.
  Finally, we demonstrate the effectiveness of IRGP-UCB through extensive experiments.
\end{abstract}



\section{Introduction}
\label{sec:intro}

\emph{Bayesian optimization} (BO) \citep{Mockus1978-Application} has become a widely-used framework for expensive black-box optimization problems.
To reduce the number of function evaluations, BO sequentially observes the noisy function value using an acquisition function computed from a Bayesian model.
BO has been applied to many different fields, including automatic machine learning \citep{Snoek2012-Practical,kandasamy2018-neural}, materials informatics \citep{ueno2016combo}, and drug design \citep{korovina2020-chembo,griffiths2020-constrained}.

Theoretical guarantees for BO have also been studied extensively \citep[e.g., ][]{Srinivas2010-Gaussian, Russo2014-learning}.
The Gaussian process upper confidence bound (GP-UCB) \citep{Srinivas2010-Gaussian}, which achieves a strong theoretical guarantee for cumulative regret bounds, is a seminal work in this field.
Based on this versatile framework for the analysis, while maintaining the theoretical guarantee, GP-UCB has been extended to various problems, which include multi-objective BO \citep{paria2020-flexible,Zuluga2016-epsilon}, multi-fidelity BO \citep{Kandasamy2016-Gaussian,Kandasamy2017-Multi}, parallel BO \citep{Contal2013-Parallel,Desautels2014-Parallelizing}, high-dimensional BO \citep{kandasamy2015-high}, and cascade BO \citep{Kusakawa2022-bayesian}.

However, the choice of a \emph{confidence parameter} $\beta_t$, which controls the exploitation-exploration trade-off, is a practical challenge for GP-UCB-based methods, where $t$ is the iteration of BO.
Since the theoretical choice of $\beta_t$ is considerably large and increases with $t$, particularly in the later iterations, GP-UCB with the theoretical $\beta_t$ causes over-exploration, which reduces the optimization performance.
Therefore, in practice, $\beta_t$ is often specified manually using some heuristics, which strongly affects the performance of GP-UCB.
Since this is a common problem with GP-UCB-based methods, mitigating the effect of the manual specification of $\beta_t$ is an important task.

To resolve this issue, \citet{berk2021-randomized} have proposed a \emph{randomized} GP-UCB (RGP-UCB), which uses a randomized confidence parameter $\zeta_t$, which follows the Gamma distribution.
Their experimental results demonstrated that RGP-UCB performed better than the standard GP-UCB.
Although \citet{berk2021-randomized} provided the regret analysis, it appears to contain some technical issues, including an unknown convergence rate (See Appendix~\ref{app:issues_RGPUCB} for details).
%
%
Therefore, the theoretical justification of RGP-UCB is still an open problem.

\begin{table*}[ht]
    \centering
    \begin{tabular}{c|c|c|c}
         & GP-UCB & RGP-UCB & IRGP-UCB (proposed) \\ \hline
         BCR (discrete) & $O(\sqrt{T \gamma_T \log T})$ & $O(\sqrt{T \gamma_T \log T})$ & * $O(\sqrt{T \gamma_T})$ \\
         BCR (continuous) & $O(\sqrt{T \gamma_T \log T})$ & $O(\sqrt{T \gamma_T \log T})$ & $O(\sqrt{T \gamma_T \log T})$ \\
         Sufficient condition of BSR $< \eta$ & $\sqrt{ \gamma_T \log T / T} \lesssim \eta$ &  $\sqrt{\gamma_T \log T / T} \lesssim \eta$ &  * $\sqrt{\gamma_T  / T} \lesssim \eta / \sqrt{- \log \eta}$
    \end{tabular}
    \caption{
        Summary of Bayesian regret of GP-UCB-based algorithms.
        The first and second rows show the BCR bounds for discrete and continuous input domains, respectively, where $\gamma_T$ is the maximum information gain defined in Section~\ref{sec:background}.
        The third row shows the sufficient conditions to achieve that BSR is lower than the predefined accuracy $\eta \in (0, 1)$ for both continuous and discrete input domains, where $\lesssim$ represents an inequality ignoring factors except for $T$ and $\eta$.
        Star means better bounds than known results with respect to $T$.
        Note that the results in the right two columns are derived in this paper.
        }
    \label{tab:regret_summary}
\end{table*}

This study demonstrates a generalized regret bound of RGP-UCB, which holds for a wider class of distributions for $\zeta_t$ in the Bayesian setting, where an objective function $f$ is assumed to be a sample path from the GP.
However, this analysis requires an increase in $\zeta_t$, as $\EE [\zeta_t] \approx \beta_t$, which implies the over-exploration cannot be avoided.
Furthermore, this generalization does not provide us the choice of the distribution of $\zeta_t$.
%

The aforementioned analyses motivated us to propose an improved RGP-UCB (IRGP-UCB), in which $\zeta_t$ follows a \emph{two-parameter exponential distribution}.
First, we show the sub-linear Bayesian cumulative regret (BCR) bounds of IRGP-UCB in the Bayesian setting for both finite and infinite input domains.
In particular, when the input domain is finite, the BCR for IRGP-UCB achieves the better convergence rate, which shows a $O(\sqrt{\log T})$ multiplicative factor improvement from the known bounds, where $T$ is the entire time horizon.
%
%
Furthermore, IRGP-UCB achieves a tighter convergence rate with respect to $T$ for the Bayesian simple regret (BSR) than existing analyses for both finite and infinite domains.
More importantly, these analyses also reveal that the increase in $\zeta_t$ for IRGP-UCB is unnecessary.
Therefore, the over-exploration of the original GP-UCB is theoretically alleviated via randomization using the two-parameter exponential distribution.
%
%
%
A key finding in our proof is a direct upper bound of $\EE[\max f(\*x)]$.
We show that $\EE[\max f(\*x)]$ can be bounded by the expectation of the UCB with our random non-increasing confidence parameter, which enables us to derive tighter Bayesian regret bounds than known results.

Our main contributions are summarized as follows:
\begin{itemize}
    \item We provide the Bayesian regret bounds for RGP-UCB, which holds for a wider class of distributions for $\zeta_t$ and achieves the same convergence rate as GP-UCB.
    \item 
        We propose yet another randomized variant of GP-UCB called IRGP-UCB, which sets the confidence parameter $\zeta_t$ using the two-parameter exponential distribution. 
        A notable advantage of IRGP-UCB is that increases in the confidence parameter in proportion to $t$ are unnecessary. 
    \item 
        We show Bayesian regret bounds for IRGP-UCB, which achieves the better convergence rate, in which $O(\sqrt{\log T})$ multiplicative factor is improved from known results.
        This result suggests that the over-exploration of GP-UCB is theoretically alleviated.
    \item We provide the upper bound for $\EE[\max f(\*x)]$ along the way of the proof, which enables us to improve the Bayesian regret bounds.
\end{itemize}
The theoretical results are summarized in Table~\ref{tab:regret_summary}.
%
%
Finally, we demonstrate the effectiveness of IRGP-UCB through a wide range of experiments.

\subsection{Related Work}
\label{sec:related_work}

This study considers BO with the Bayesian setting, where the objective function $f$ is assumed to be a sample path from the GP.
Various BO methods have been developed in the literature, for example, expected improvement (EI) \citep{Mockus1978-Application}, entropy search (ES) \citep{Henning2012-Entropy}, and predictive entropy search (PES) \citep{Hernandez2014-Predictive}.
Although the regret analysis of EI for the noiseless and frequentist setting, in which $f$ is an element of reproducing kernel Hilbert space, is provided in \citep{bull2011convergence}, the Bayesian setting has not been considered.
Further, although the practical performance of ES and PES has been shown repeatedly, their regret analysis is an open problem.

GP-UCB is one of the prominent studies for theoretically guaranteed BO methods.
\citet{Srinivas2010-Gaussian} showed the high probability bound for cumulative regret.
Although Bayesian regret analysis for GP-UCB has not been shown explicitly, \citet{paria2020-flexible} have shown the BCR bound for a multi-objective extension of GP-UCB, which contains that of the original single-objective GP-UCB as the special case.
For completeness, we show the BCR bound of GP-UCB in Section~\ref{sec:GPUCB}.
Although many studies \citep[e.g., ][]{Srinivas2010-Gaussian,Chowdhury2017-on,janz2020-bandit} considered the frequentist setting, this study concentrates on the Bayesian setting.
%

\citet{berk2021-randomized} attempted to alleviate the GP-UCB hyperparameter tuning issue through randomization using the Gamma distribution.
Note that our result shown in Theorem~\ref{theo:BCR_RGPUCB_informal} differs from \citep{berk2021-randomized}.
We believe that these differences come from several technical issues in the proof in \citep{berk2021-randomized} (See Appendix~\ref{app:issues_RGPUCB} for details).
In addition, the final convergence rate in \citep{berk2021-randomized} includes the variables whose convergence rate is not proved.
In contrast, our Theorem~\ref{theo:BCR_RGPUCB_informal} fully clarifies the convergence rate without those unproved variables.
Moreover, Theorem~\ref{theo:BCR_RGPUCB_informal} is generalized to a wider class of distributions for confidence parameters that contain the Gamma distribution.
In addition, the two-parameter exponential distribution used in IRGP-UCB is not considered in \citep{berk2021-randomized}.

Another standard BO method with a theoretical guarantee is Thompson sampling (TS) \citep{Russo2014-learning,Kandasamy2018-Parallelised}.
TS achieves the sub-linear BCR bounds by sequentially observing the optimal point of the GP's posterior sample path.
Although TS does not require any hyperparameter, TS is often deteriorated by over-exploration, as discussed in \citep{Shahriari2016-Taking}.
%

\citet{Wang2016-Optimization} have shown the regret analysis of the GP estimation (GP-EST) algorithm, which can be interpreted as GP-UCB with the confidence parameter defined using $\hat{m}$, an estimator of $\EE[\max f(\*x)]$.
Their analysis requires an assumption $\hat{m} \geq \EE[\max f(\*x)]$, whose sufficient condition is provided in \citep[][Corollary~3.5]{Wang2016-Optimization}.
However, this sufficient condition does not typically hold, as discussed immediately after the corollary in \citep{Wang2016-Optimization}.
Furthermore, the final convergence rate contains $\hat{m}$ itself, whose convergence rate is not clarified.
In contrast, our Lemma~\ref{lem:bound_RGPUCB} shows the bound for $\EE[\max f(\*x)]$ under common regularity conditions.
\citet{Wang2017-Max} have shown the regret analysis of max-value entropy search (MES).
However, it is pointed out that their proof contains several technical issues \citep{takeno2022-sequential}.

\section{Background}
\label{sec:background}

\subsection{Bayesian Optimization}

We consider an optimization problem $\*x^* = \argmax_{\*x \in \cX} f(\*x)$, where $f$ is an unknown expensive-to-evaluate objective function, $\cX \subset \RR^d$ is an input domain, and $d$ is an input dimension.
BO sequentially observes the noisy function value aiming to minimize the number of function evaluations.
Thus, at each iteration $t$, we can query $\*x_t$ and obtain $y_t = f(\*x_t) + \epsilon_t$, where $\epsilon_t \sim \cN(0, \sigma^2)$ is i.i.d. Gaussian noise with a positive variance $\sigma^2 > 0$.

We assume that $f$ is a sample path from a GP \citep{Rasmussen2005-Gaussian} with a zero mean and a stationary kernel function $k: \cX \times \cX \mapsto \RR$ denoted as $f \sim \cG \cP(0, k)$.
At each iteration $t$, a dataset $\cD_{t-1} \coloneqq \{ (\*x_i, y_i) \}_{i=1}^{t-1}$ is already obtained from the nature of BO.
Then, the posterior distribution $p(f \mid \cD_{t-1})$ is a GP again.
The posterior mean and variance of $f(\*x)$ are derived as follows:
\begin{align*}
    \mu_{t-1}(\*x) &= \*k_{t-1}(\*x)^\top \bigl(\*K + \sigma^2 \*I_{t-1} \bigr)^{-1} \*y_{t-1}, \\
    \sigma_{t-1}^2 (\*x) &= k(\*x, \*x) - \*k_{t-1}(\*x) ^\top \bigl(\*K + \sigma^2 \*I_{t-1} \bigr)^{-1} \*k_{t-1}(\*x),
\end{align*}
where $\*k_{t-1}(\*x) \coloneqq \bigl( k(\*x, \*x_1), \dots, k(\*x, \*x_{t-1}) \bigr)^\top \in \RR^{t-1}$, $\*K \in \RR^{(t-1)\times (t-1)}$ is the kernel matrix whose $(i, j)$-element is $k(\*x_i, \*x_j)$, $\*I_{t-1} \in \RR^{(t-1)\times (t-1)}$ is the identity matrix, and $\*y_{t-1} \coloneqq (y_1, \dots, y_{t-1})^\top \in \RR^{t-1}$.
Hereafter, we denote that the probability density function (PDF) $p(\cdot \mid \cD_{t-1}) = p_t (\cdot)$, the probability $\Pr (\cdot \mid \cD_{t-1}) = \myPr_t (\cdot) $, and the expectation $\EE[\cdot \mid \cD_{t-1}] = \EE_t[\cdot]$ for brevity.

\subsection{Preliminaries for Regret Analysis}

When $\cX$ is infinite (continuous), the following regularity assumption is used in most analyses \citep[e.g., ][]{Srinivas2010-Gaussian,Kandasamy2018-Parallelised,paria2020-flexible}:
\begin{assumption}
    Let $\cX \subset [0, r]^d$ be a compact and convex set, where $r > 0$.
    Assume that the kernel $k$ satisfies the following condition on the derivatives of a sample path $f$.
    There exists the constants $a, b > 0$ such that,
    \begin{align*}
        \Pr \left( \sup_{\*x \in \cX} \left| \frac{\partial f}{\partial \*x_j} \right| > L \right) \leq a \exp \left( - \left(\frac{L}{b}\right)^2 \right),\text{ for } j \in [d],
    \end{align*}
    where $[d] = \{1, \dots, d\}$.
    \label{assump:continuous_X}
\end{assumption}
Our analysis also requires this assumption.

The convergence rates of regret bounds are characterized by \emph{maximum information gain} (MIG) \citep{Srinivas2010-Gaussian,vakili2021-information}.
MIG $\gamma_T$ is defined as follows:
\begin{definition}[Maximum information gain]
    Let $f \sim \cG \cP (0, k)$.
    Let $A = \{ \*a_i \}_{i=1}^T \subset \cX$.
    Let $\*f_A = \bigl(f(\*a_i) \bigr)_{i=1}^T$, $\*\epsilon_A = \bigl(\epsilon_i \bigr)_{i=1}^T$, where $\forall i, \epsilon_i \sim \cN(0, \sigma^2)$, and $\*y_A = \*f_A + \*\epsilon_A \in \RR^T$.
    Then, MIG $\gamma_T$ is defined as follows:
    \begin{align*}
        \gamma_T \coloneqq \argmax_{A \subset \cX; |A| = T} I(\*y_A ; \*f_A),
    \end{align*}
    where $I$ is the Shanon mutual information.
\end{definition}
MIG is known to be sub-linear for commonly used kernel functions, e.g., $\gamma_T = O\bigl( (\log T)^{d+1} \bigr)$ for RBF kernels and $\gamma_T = O\bigl( T^{\frac{d}{2\nu + d}} (\log T)^{\frac{2\nu}{2\nu + d}} \bigr)$ for Mat\`{e}rn-$\nu$ kernels \citep{Srinivas2010-Gaussian,vakili2021-information}.

\subsection{Bayesian Regret Analysis}

In this paper, we evaluate the performance of BO methods by \emph{Bayesian regret} \citep{Russo2014-learning,Kandasamy2018-Parallelised,paria2020-flexible}.
The BSR\footnote{This definition of BSR only requires that some observed input, which may be unknown due to the noise, achieves low regret. We define BSR using the input recommended by the algorithm and show its convergence in Appendix~\ref{app:BSR_recommendation}.} and BCR are defined as follows:
\begin{align}
    {\rm BSR}_T \coloneqq \EE\bigl[ f(\*x^*) - \max_{t \leq T} f(\*x_t) \bigr], \label{eq:BSR} \\
    {\rm BCR}_T \coloneqq \EE\left[\sum_{t=1}^T f(\*x^*) - f(\*x_t) \right], \label{eq:BCR}
\end{align}
where the expectation is taken with all randomness, including $f, \epsilon_t$, and the randomness of BO algorithms.
We discuss the convergence of BSR by analyzing the BCR.
That is, if BCR is sub-linear, BSR converges to zero since ${\rm BSR}_T \leq \frac{1}{T} {\rm BCR}_T$.
Furthermore, we evaluate a required number of function evaluations to achieve ${\rm BSR}_T \leq \eta$, where $\eta > 0$ represents the desired accuracy of a solution.

According to \citep{Russo2014-learning}, Bayesian regret bounds directly provide high probability bounds.
Assume ${\rm BCR}_T = O(h(T))$ with some non-negative function $h$.
Then, the direct consequence of Markov's inequality implies that $\sum_{t=1}^T f(\*x^*) - f(\*x_t) = O \left( h(T) / \delta \right)$ holds with probability $1 - \delta$, where $\delta \in (0, 1)$.
Thus, the improvement of the convergence rate of BCR for $T$ shows that of high-probability bound although the rate for $\delta$ is worse compared to $\log (1/\delta)$ \citep{Srinivas2010-Gaussian}.

The existing Bayesian regret analyses typically use the technique called \emph{regret decomposition} \citep{Russo2014-learning}.
This approach decomposes BCR as follows:
\begin{align}
    {\rm BCR}_T = \sum_{t=1}^T \EE\left[ f(\*x^*) - u_t(\*x_t) \right] + \EE\left[ u_t(\*x_t) - f(\*x_t) \right],
    \label{eq:regret_depomsition}
\end{align}
where $u_t(\*x) \coloneqq \mu_{t-1}(\*x) + \beta^{1/2}_t \sigma_{t-1}(\*x)$.
If we use fixed constant for $\{\beta_t\}_{t \geq 1}$ (e.g., $\forall t\geq 1, \beta_t = 2$), bounding the first term $\sum_{t=1}^T \EE\left[ f(\*x_*) - u_t(\*x_t) \right]$ is difficult.
Thus, in existing analyses, $\{\beta_t\}_{t \geq 1}$ is scheduled so that $\beta_t \propto \log t$.
Since $\beta_t$ remains in the upper bound, the regret decomposition using $u_t(\*x)$ deteriorates the final convergence rate, as we will show an example of GP-UCB in Theorem~\ref{theo:BCR_GPUCB_informal}.
IRGP-UCB avoids this problem by randomizing the confidence parameters, as shown in Section~\ref{sec:theorem}.

\subsection{GP-UCB}
\label{sec:GPUCB}

The GP-UCB \citep{Srinivas2010-Gaussian} selects the next evaluation by maximizing UCB as follows:
\begin{align*}
    \*x_t = \argmax_{\*x \in \cX} \mu_{t-1}(\*x) + \beta^{1/2}_t \sigma_{t-1}(\*x).
\end{align*}
Although BCR bound for the standard GP-UCB has not been shown explicitly, sub-linear regret bounds of GP-UCB can be shown in nearly the same way as \citep{paria2020-flexible}:
\begin{theorem}[Informal: BCR of GP-UCB]
    Suppose that $f \sim \cG \cP(0, k)$, where $k(\*x, \*x) = 1$.
    Assume that $\cX$ is finite or Assumption~\ref{assump:continuous_X} holds.
    Then, by running GP-UCB with $\beta_t \propto \log(t)$, BCR can be bounded as follows: 
    \begin{align}
        {\rm BCR}_T = O(\sqrt{T \beta_T \gamma_T}),
    \end{align}
    which implies sub-linearity when $\gamma_T = o(T / \log T)$.
    \label{theo:BCR_GPUCB_informal}
\end{theorem}
See Appendix~\ref{app:GP_UCB_proof} for the proof\footnote{It is worth noting that the convergence rate for $a$ in Assumption~\ref{assump:continuous_X} is tightened in our proof compared with the proofs in \citep{paria2020-flexible,Kandasamy2018-Parallelised}. This is applied to all subsequent regret analyses for a continuous domain. See details for Appendixes~\ref{app:GP_UCB_proof} and \ref{app:lemmas}.}.
In the theorem, the confidence parameter $\beta_t$ must be scheduled as $\beta_t \propto \log(t)$.
%

\section{Algorithm}
\label{sec:algorithm}

Algorithm \ref{alg:RGPUCB} shows a simple procedure of IRGP-UCB.
%
The difference from the original GP-UCB is that $\zeta_t$ is a random variable, not a constant.
Using generated $\zeta_t$, the next evaluation is chosen as follows:
\begin{align*}
    \*x_t = \argmax_{\*x \in \cX} \mu_{t-1}(\*x) + \zeta^{1/2}_t \sigma_{t-1}(\*x).
\end{align*}
Our choice of the distribution of $\zeta_t$ is a \emph{two-parameter exponential distribution} \citep[e.g., ][]{beg1980estimation,lam1994estimation}, whose PDF is written as follows:
\begin{align*}
    p(\zeta; s, \lambda) = 
    \begin{cases}
    \lambda \exp\bigl( \lambda (\zeta - s) \bigr) & \text{if } \zeta \geq s, \\
    0 & \text{otherwise},
    \end{cases}
\end{align*}
where $\lambda$ is a rate parameter.
This can be seen as a distribution of the sum of $s$ and $Z \sim {\rm Exp} (\lambda)$.
Thus, the sampling from the two-parameter exponential distribution can be easily performed using an exponential distribution as follows:
\begin{align}
    \zeta_t \gets s + Z, \text{ where } Z \sim {\rm Exp} (\lambda).
    \label{eq:sampling_beta}
\end{align}
The theoretical choice of $s$ and $\lambda$ will be shown in Section~\ref{sec:theorem}.

\begin{algorithm}[!t]
    \caption{IRGP-UCB}\label{alg:RGPUCB}
    \begin{algorithmic}[1]
        \Require Input space $\cX$, Parameters $s$ and $\lambda$ for $\zeta_t$, GP prior $\mu=0$ and $k$
        \State $\cD_{0} \gets \emptyset $
        \For{$t = 1, \dots$}
            \State Fit GP to $\cD_{t-1}$
            \State Generate $\zeta_t$ by Eq.~\eqref{eq:sampling_beta}
            \State $\*x_t \gets \argmax_{\*x \in \cX} \mu_{t-1}(\*x) + \zeta_t^{1/2} \sigma_{t-1}(\*x)$
            \State Observe $y_t = f(\*x_t) + \epsilon_t$ and $\cD_{t} \gets \cD_{t-1} \cup (\*x_t, y_t)$
        \EndFor
    \end{algorithmic}
\end{algorithm}

\section{Regret Analysis}
\label{sec:theorem}

First, we show the Bayesian regret analysis for a general RGP-UCB, which is not restricted to the two-parameter exponential distribution.
Next, we show the tighter Bayesian regret bounds for IRGP-UCB.

\subsection{Regret Bound for General RGP-UCB}

Here, we provide a generalized theoretical analysis for RGP-UCB, which holds for a wider class of distributions.
\begin{theorem}[Informal: BCR of RGP-UCB]
    Suppose that $f \sim \cG \cP(0, k)$, where $k(\*x, \*x) = 1$.
    Assume that $\cX$ is finite or Assumption~\ref{assump:continuous_X} holds.
    Let $\{\zeta_t\}_{t \geq 1}$ be a sequence of non-negative random variables, which satisfies $\EE[\zeta_t] = O(\log t)$ and $\EE[\exp ( - \zeta_t / 2)] = O (1 / t^2)$.
    By running RGP-UCB with $\zeta_t$, BCR can be bounded as follows: 
    \begin{align*}
        {\rm BCR}_T = O(\sqrt{T \gamma_T \EE[\zeta_T]}),
    \end{align*}
    which implies sub-linearity when $\gamma_T = o(T / \log T)$.
    \label{theo:BCR_RGPUCB_informal}
\end{theorem}
See Appendix~\ref{app:proof_general_RGPUCB} for the proof, in which we also rectify the technical issues in \citep{berk2021-randomized} listed in Appendix~\ref{app:issues_RGPUCB}.
A notable difference between Theorem~\ref{theo:BCR_RGPUCB_informal} and \citep[][Theorem~3]{berk2021-randomized} is that Theorem~\ref{theo:BCR_RGPUCB_informal} is applicable to a wider class of distributions of $\zeta_t$, not only Gamma distribution proposed in \citep{berk2021-randomized}.
Roughly speaking, Theorem~\ref{theo:BCR_RGPUCB_informal} only requires that the distribution has the parameters that can control the distribution so that $\EE[\zeta_t] = O(\log t)$ and $\EE[\exp ( - \zeta_t / 2)] = O (1 / t^2)$.
This condition can be satisfied by many distributions, such as Gamma, two-parameter exponential, and truncated normal distributions, as shown in Appendix~\ref{app:examples_RGPUCB_dist}.
%
%
See Appendix~\ref{app:examples_RGPUCB_dist} for a detailed discussion.

Although Theorem~\ref{theo:BCR_RGPUCB_informal} achieves sub-linear BCR bounds, the implication of Theorem~\ref{theo:BCR_RGPUCB_informal} is restricted.
In the proof, $\zeta_t$ must behave similar to $\beta_t$, that is, $\EE[\zeta_t] \propto \beta_t$ and $\EE[\exp ( - \zeta_t / 2)] \propto \exp ( - \beta_t / 2)$.
Therefore, based on this theorem, over-exploration of the GP-UCB cannot be avoided.
Furthermore, this theorem does not specify a distribution, that is suitable for the randomization of GP-UCB.
%

\subsection{Regret Bounds for IRGP-UCB}

First, for completeness, we provide a modified version of the basic lemma derived in \citep{Srinivas2010-Gaussian}:
\begin{lemma}
    Suppose that $f$ is a sample path from a GP with zero mean and a stationary kernel $k$ and $\cX$ is finite.
    Pick $\delta \in (0, 1)$ and $t \geq 1$.
    Then, for any given $\cD_{t-1}$,
    \begin{align*}
        \myPr_t \left( f(\*x) \leq \mu_{t-1}(\*x) + \beta^{1/2}_{\delta} \sigma_{t-1}(\*x), \forall \*x \in \cX \right)
        \geq 1 - \delta,
    \end{align*}
    where $\beta_{\delta} = 2 \log (|\cX| /( 2 \delta))$.
    \label{lem:bound_srinivas}
\end{lemma}
See Appendix~\ref{app:proof_srinivas} for the proof.
This lemma differs slightly from \citep[][Lemma~5.1]{Srinivas2010-Gaussian}, since, in Lemma~\ref{lem:bound_srinivas}, the iteration $t$ is fixed, and $\beta_{\delta}$ does not depend on $t$.

Based on Lemma~\ref{lem:bound_srinivas}, we show the following key lemma to show the improved convergence rate:
\begin{lemma}
    Let $f \sim \cG \cP (0, k)$, where $k$ is a stationary kernel and $k(\*x, \*x) = 1$, and $\cX$ be finite.
    Assume that $\zeta$ follows a two-parameter exponential distribution with $s = 2 \log (|\cX| / 2)$ and $\lambda = 1/2$.
    Then, the following inequality holds:
    \begin{align*}
        \EE[f(\*x^*)] \leq \EE \left[\max_{\*x \in \cX} \mu_{t-1}(\*x) + \zeta^{1/2}_t \sigma_{t-1}(\*x) \right],
    \end{align*}
    for all $t \geq 1$.
    \label{lem:bound_RGPUCB}
\end{lemma}
\begin{proof}
    We here show the short proof of Lemma~\ref{lem:bound_RGPUCB} although detailed proof is shown in Appendix~\ref{app:proof_lemma_IRGPUCB}.
    From the tower property of the expectation, it suffices to show the following inequality:
    \begin{align*}
        \EE_t[f(\*x^*)] \leq \EE_t \left[\max_{\*x \in \cX} \mu_{t-1}(\*x) + \zeta^{1/2}_t \sigma_{t-1}(\*x) \right].
    \end{align*}
    Since this inequality only considers the expectation given $\cD_{t-1}$, we can fix $\cD_{t-1}$ in the proof.
    Furthermore, from Lemma~\ref{lem:bound_srinivas}, we can obtain the following inequality:
    \begin{align*}
        F_t^{-1}(1 - \delta) \leq \max_{\*x \in \cX} \mu_{t-1}(\*x) + \beta^{1/2}_{\delta} \sigma_{t-1}(\*x),
    \end{align*}
    where $F_t (\cdot) \coloneqq \myPr_t\left( f(\*x_*) < \cdot \right)$ is a cumulative distribution function of $f(\*x_*)$, and $F_t^{-1}$ is its inverse function.
    Then, substituting $U \sim {\rm Uni}(0, 1)$ into $\delta$ and taking the expectation, we obtain the following inequality:
    \begin{align*}
        \EE_t \left[ f(\*x_*) \right] \leq \EE_U \left[ \max_{\*x \in \cX} \mu_{t-1}(\*x) + \beta^{1/2}_{U} \sigma_{t-1}(\*x) \right],
    \end{align*}
    which can be derived in a similar way to inverse transform sampling.
    Hence, $\beta^{1/2}_{U} = 2 \log (|\cX| / 2) - 2 \log(U)$ results in a random variable, which follows the two-parameter exponential distribution.
    Consequently, we can obtain the desired bound.
\end{proof}

%
%

Lemma~\ref{lem:bound_RGPUCB} shows a direct upper bound of the expectation of $f(\*x^*)$.
A remarkable consequence of Lemma~\ref{lem:bound_RGPUCB} is an upper bound of BCR as follows:
\begin{align*}
    {\rm BCR}_T 
    &= \sum_{t=1}^T \EE\left[ f(\*x^*) - v_t(\*x_t) \right] + \EE\left[ v_t(\*x_t) - f(\*x_t) \right] \\
    &\leq \sum_{t=1}^T \EE\left[ v_t(\*x_t) - f(\*x_t) \right],
\end{align*}
where $v_t(\*x) \coloneqq \mu_{t-1}(\*x) + \zeta^{1/2}_t \sigma_{t-1}(\*x)$.
This upper bound eliminates the first term, which cannot be bounded without increasing the confidence parameter in the conventional regret decomposition Eq.~\eqref{eq:regret_depomsition}. 
%
%
Importantly, $\EE[\zeta_t] = 2 + s$ is a constant since $s$ and $\lambda$ for $\zeta$ does not depend on $t$ in Lemma~\ref{lem:bound_RGPUCB}.
Thus, using Lemma~\ref{lem:bound_RGPUCB}, we can obtain the tighter BCR bound for the finite input domain:
\begin{theorem}[BCR bound for finite domain]
    Let $f \sim \cG \cP (0, k)$, where $k$ is a stationary kernel and $k(\*x, \*x) = 1$, and $\cX$ be finite.
    Assume that $\zeta_t$ follows a two-parameter exponential distribution with $s = 2 \log (|\cX| / 2)$ and $\lambda = 1/2$ for any $t \geq 1$.
    Then, by running IRGP-UCB with $\zeta_t$, BCR can be bounded as follows: 
    \begin{align*}
        {\rm BCR}_T \leq \sqrt{C_1 C_2 T \gamma_T},
    \end{align*}
    where $C_1 \coloneqq 2 / \log(1 + \sigma^{-2})$ and $C_2 \coloneqq 2 + s$. 
    \label{theo:BCR_IRGPUCB_discrete}
\end{theorem}
See Appendix~\ref{app:proof_discrete} for detailed proof.

Theorem~\ref{theo:BCR_IRGPUCB_discrete} has two important implications.
First, Theorem~\ref{theo:BCR_IRGPUCB_discrete} shows the convergence rate $O(\sqrt{T \gamma_T})$, in which multiplicative $\sqrt{\log T}$ factor is improved compared with $O(\sqrt{T \gamma_T \log T})$ achieved by GP-UCB \citep{Srinivas2010-Gaussian}, RGP-UCB, and TS \citep{Russo2014-learning}.
%
%
Second, more importantly, IRGP-UCB does not need to schedule the parameters of $\zeta_t$, i.e., $s$ and $\lambda$, in contrast to GP-UCB and RGP-UCB.
Therefore, through randomization, IRGP-UCB essentially alleviates the problem that the well-known $\beta_t \propto \log(t)$ strategy results in a practically too large confidence parameter.
Further, over-exploration in the later iterations can be avoided.
The IRGP-UCB is the first GP-UCB-based method that enjoys the above technical and practical benefits.

On the other hand, note that the dependence $\sqrt{\log |\cX|}$ remains as with the prior works \citep{Srinivas2010-Gaussian,Russo2014-learning,Kandasamy2018-Parallelised,paria2020-flexible}.
In BO, we are usually interested in the case that the total number of iterations $T < |\cX|$.
Thus, $\sqrt{\log |\cX|}$ is dominant compared with $\sqrt{\log T}$ factor improvement.

It is worth noting that our key lemma (Lemma~\ref{lem:bound_RGPUCB}) mainly requires Lemma~\ref{lem:bound_srinivas} only, which is expected to be satisfied in a wide range of exploration problems in which a GP is used as a surrogate model. 
Therefore, we conjecture that the same proof technique can be applicable to more advanced problem settings, such as multi-objective BO \citep{paria2020-flexible}, for which further analyses are important future directions of our results.

Next, we show the BCR bound for the infinite (continuous) domain:
\begin{theorem}[BCR bound for infinite domain]
    Let $f \sim \cG \cP (0, k)$, where $k$ is a stationary kernel, $k(\*x, \*x) = 1$, and Assumption~\ref{assump:continuous_X} holds.
    Assume that $\zeta_t$ follows a two-parameter exponential distribution with $s_t = 2d \log(bdr t^2 \bigl( \sqrt{\log (ad)} + \sqrt{\pi} / 2\bigr)) - 2 \log 2$ and $\lambda = 1/2$ for any $t \geq 1$.
    Then, by running IRGP-UCB, BCR can be bounded as follows: 
    \begin{align*}
        {\rm BCR}_T \leq \frac{\pi^2}{6} + \sqrt{C_1 T \gamma_T (2 + s_T)},
    \end{align*}
    where $C_1 \coloneqq 2 / \log(1 + \sigma^{-2})$. 
    \label{theo:BCR_IRGPUCB_continuous}
\end{theorem}
See Appendix~\ref{app:proof_continuous} for the proof.
Unfortunately, in this case, $\EE[\zeta_t] = O(\log t)$ and the resulting BCR bound is $O(\sqrt{T \gamma_T \log T})$.
This is because the discretization of input domain $\cX_t$, which is refined as $|\cX_t| = O(t^{2d})$, is required to obtain the BCR bound for the infinite domain.
Since we cannot avoid the dependence on $|\cX|$ also in Theorem~\ref{theo:BCR_IRGPUCB_discrete}, the resulting BCR bound in Theorem~\ref{theo:BCR_IRGPUCB_continuous} requires the term $\log(|\cX_t|) = O\bigl(2d \log(t) \bigr)$.

%
As discussed above, our proof for BCR needs to refine the discretization as $|\cX_t| = O(t^{2d})$ to bound the summation of the discretization error. 
On the other hand, if we aim to bound ${\rm BSR}_T < \eta$, the discretization error at an iteration is only required to be smaller than $\eta$. 
Therefore, the refinement of the discretization is not necessarily needed.
Hence, we can obtain the following corollaries for finite and infinite input domains, respectively:
\begin{corollary}[BSR bound for finite domain]
    Fix a required accuracy $\eta > 0$.
    Assume the same condition as in Theorem~\ref{theo:BCR_IRGPUCB_discrete}.
    Then, by running IRGP-UCB, ${\rm BSR}_T \leq \eta$ by at most $T$ function evaluations, where $T$ is the smallest positive integer satisfying the following inequality:
    \begin{align*}
        \sqrt{\frac{C_1 C_2 \gamma_T}{T}} \leq \eta,
    \end{align*}
    where $C_1 \coloneqq 2 / \log(1 + \sigma^{-2})$ and $C_2 \coloneqq 2 + s$.
    \label{coro:BSR_IRGPUCB_discrete}
\end{corollary}
\begin{corollary}[BSR bound for infinite domain]
    Fix a required accuracy $\eta > 0$.
    Assume the same condition as in Theorem~\ref{theo:BCR_IRGPUCB_continuous} except for $s_\eta = 2d \log(2bdr \bigl( \sqrt{\log (ad)} + \sqrt{\pi} / 2\bigr) / \eta) - 2 \log 2$.
    Then, by running IRGP-UCB, ${\rm BSR}_T \leq \eta$ by at most $T$ function evaluations, where $T$ is the smallest positive integer satisfying the following inequality:
    \begin{align*}
        \sqrt{\frac{C_1 (2 + s_\eta) \gamma_T}{T}} \leq \frac{\eta}{2},
    \end{align*}
    where  $C_1 \coloneqq 2 / \log(1 + \sigma^{-2})$.
    \label{coro:BSR_IRGPUCB_continuous}
\end{corollary}
See Appendixes~\ref{app:proof_discrete} and \ref{app:proof_continuous} for the proof, respectively.
These corollaries suggest that, for both finite and infinite domains, an arbitrary accuracy solution can be obtained using finite function evaluations, whose rate is shown by the inequalities.
In Corollary~\ref{coro:BSR_IRGPUCB_continuous}, $s_\eta$ depends on $\eta$ instead of $t$.
Thus, in both corollaries, the convergence rate of the left-hand side with respect to $T$ is $O(\sqrt{\gamma_T / T})$, which is improved compared with $O(\sqrt{\gamma_T \log T / T})$ achieved by GP-UCB, RGP-UCB, and TS.
This faster convergence rate is an essential improvement derived based on Lemma~\ref{lem:bound_RGPUCB}.
Therefore, IRGP-UCB does not need to schedule $\zeta_t$ if we require optimization rather than cumulative regret minimization.

\section{Experiments}
\label{sec:experiment}

We demonstrate the experimental results on synthetic and benchmark functions and the materials dataset provided in \citep{liang2021benchmarking}.
As a baseline, we performed EI \citep{Mockus1978-Application}, TS \citep{Russo2014-learning}, MES \citep{Wang2017-Max}, joint entropy search (JES) \citep{hvarfner2022-joint}, and GP-UCB \citep{Srinivas2010-Gaussian}.
For the posterior sampling in TS, MES, and JES, we used random Fourier features \citep{Rahimi2008-Random}.
For Monte Carlo estimation in MES and JES, we used ten samples.
We evaluate the practical performance of BO by simple regret $f(\*x^*) - \max_{t \leq T} f(\*x_t)$.

\subsection{Synthetic Function Experiment}

\begin{figure}[t]
    \centering
    \includegraphics[width=0.35\linewidth]{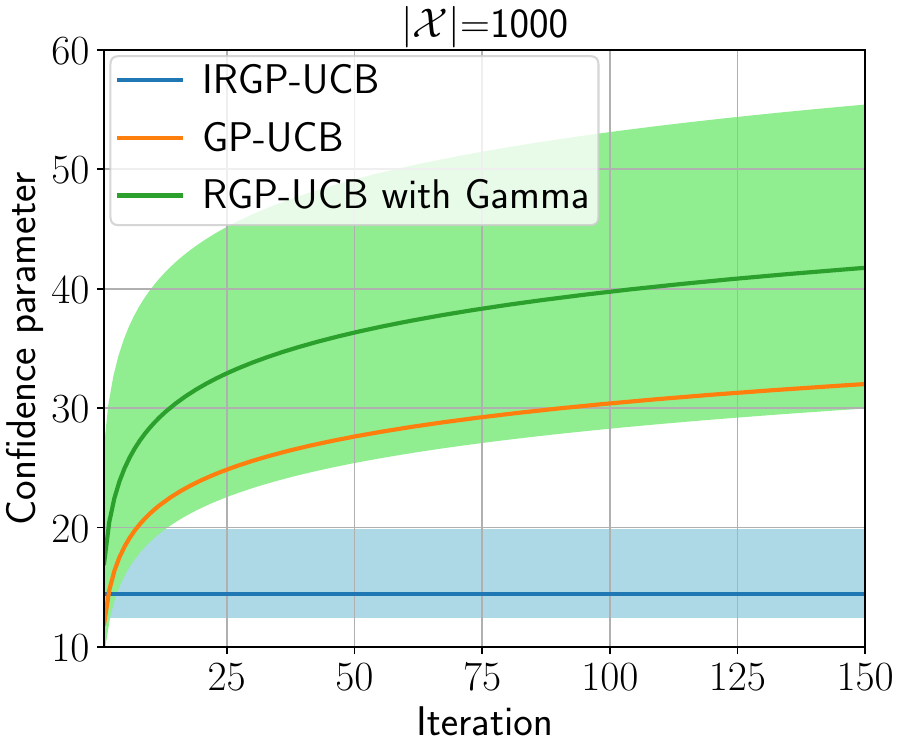}
    \caption{
        Confidence parameter of GP-UCB-based methods.
        For GP-UCB, the solid line represents $\beta_t$.
        For RGP-UCB and IRGP-UCB, the solid lines and shaded area represent the expectations $\EE[\zeta_t]$ and $95 \%$ credible intervals, respectively.
        }
    \label{fig:synthetic-confidence-param}
\end{figure}

We use synthetic functions generated from $\cG \cP (0, k)$, where $k$ is Gaussian kernel with a length scale parameter $\ell = 0.1$ and the input dimension $d = 3$.
We set the noise variance $\sigma^2 = 10^{-4}$.
The input domain consists of equally divided points in $[0, 0.9]$, i.e., $\cX = \{0, 0.1, \dots, 0.9\}^d$ and $|\cX| = 1000$.
Figure~\ref{fig:synthetic-confidence-param} shows the theoretical confidence parameters of GP-UCB, RGP-UCB with ${\rm Gamma}(\kappa_t=\log(|\cX|t^2) / \log(1.5), \theta = 1)$, and IRGP-UCB with a two-parameter exponential distribution.
We can see that the confidence parameters for GP-UCB and RGP-UCB are considerably large due to logarithmic increase, particularly in the later iterations.
In contrast, the confidence parameter of IRGP-UCB is drastically smaller than that of GP-UCB and not changed with respect to the iteration.
Therefore, we can observe that IRGP-UCB alleviates the over-exploration.

\begin{figure}[t]
    \centering
    \includegraphics[width=0.35\linewidth]{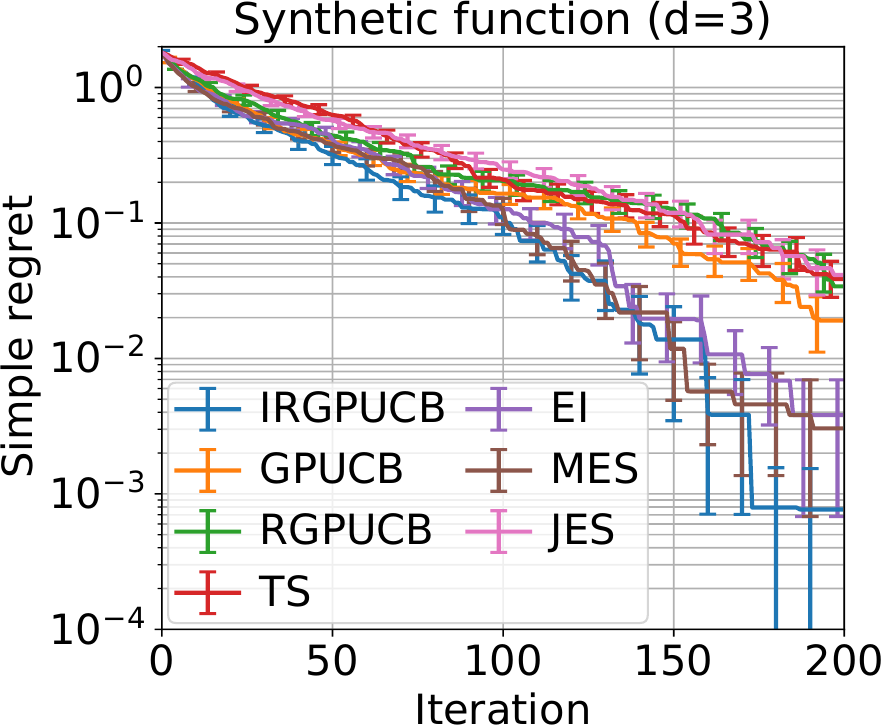}
    \caption{Average and standard errors of simple regret.}
    \label{fig:synthetic-regret}
\end{figure}

Next, we report the simple regret in Figure~\ref{fig:synthetic-regret}.
Ten synthetic functions and ten initial training datasets are randomly generated.
Thus, the average and standard error for $10 \times 10$ trials are reported.
The hyperparameters for GP are fixed to those used to generate the synthetic functions.
The confidence parameters for GP-UCB, RGP-UCB, and IRGP-UCB are set as in Figure~\ref{fig:synthetic-confidence-param}.
We can see that the regrets of GP-UCB, RGP-UCB, TS, and JES converge slowly.
We conjecture that this slow convergence came from over-exploration.
On the other hand, EI, MES, and IRGP-UCB show fast convergence.
In particular, IRGP-UCB achieves the best average in most iterations.
These results suggest that IRGP-UCB bridges the gap between theory and practice in contrast to GP-UCB and RGP-UCB.

\hfill

\subsection{Benchmark Function Experiments}


    

\begin{figure*}[!t]
    \centering
    \subfigure[Benchmark functions.]{
        \includegraphics[width=0.325\linewidth]{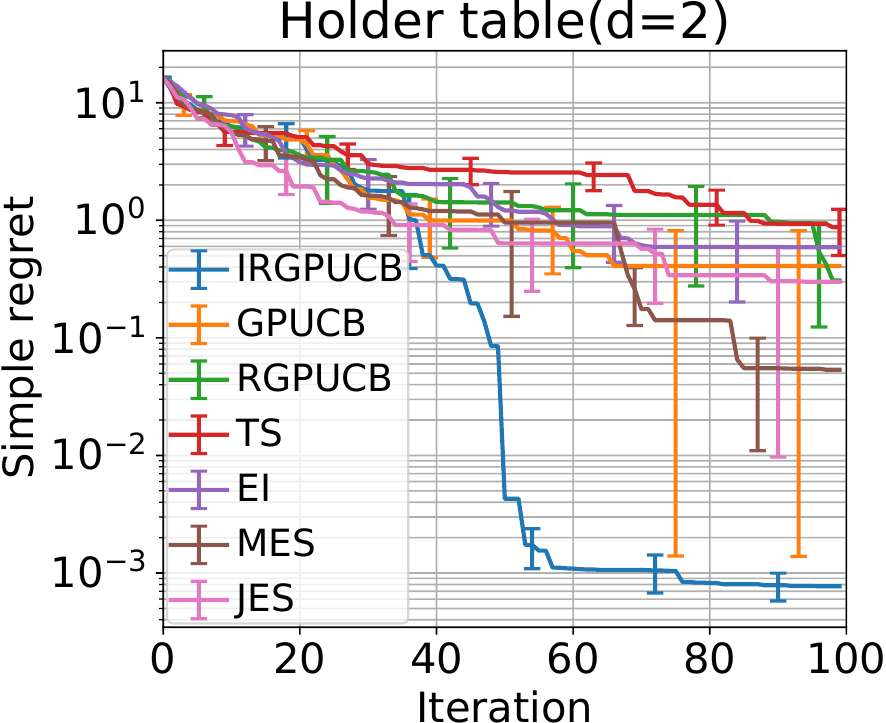}
        \includegraphics[width=0.325\linewidth]{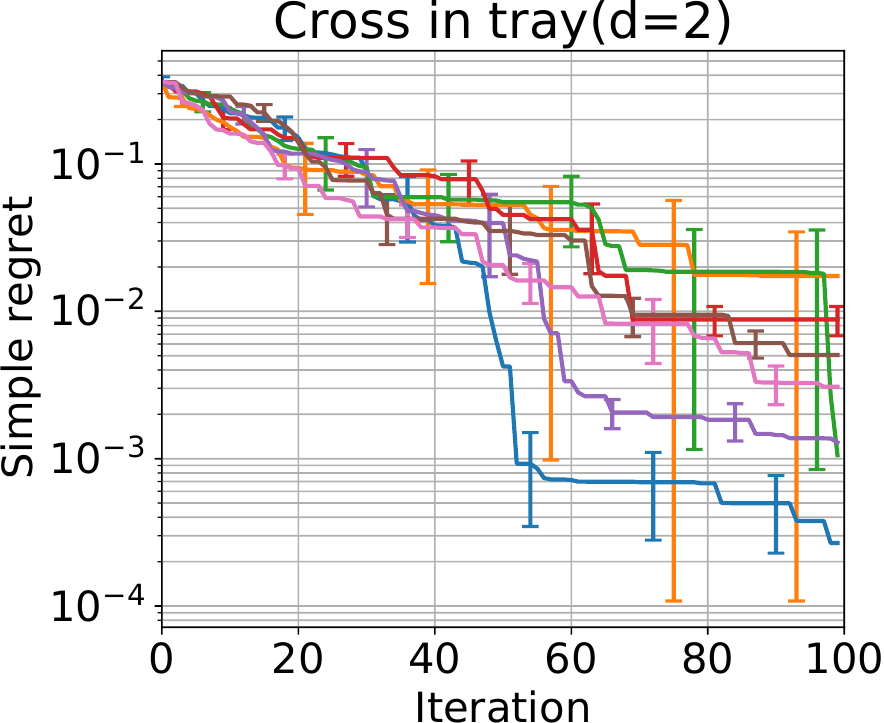}
        \includegraphics[width=0.325\linewidth]{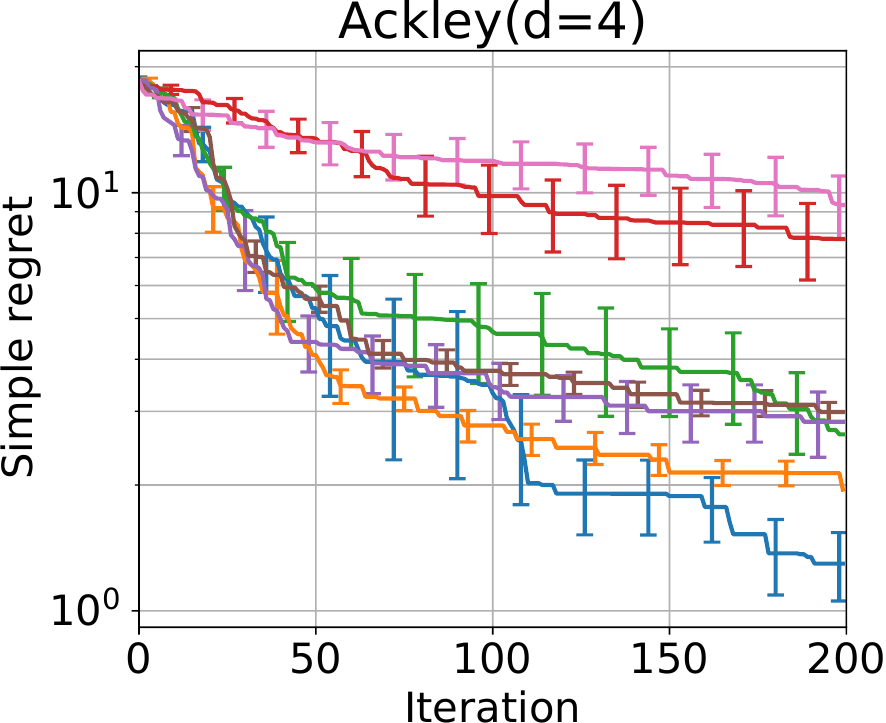}
        \label{fig:benchmark-regret}
        }
    \hfill
    \subfigure[Materials datasets.]{
        \includegraphics[width=0.325\linewidth]{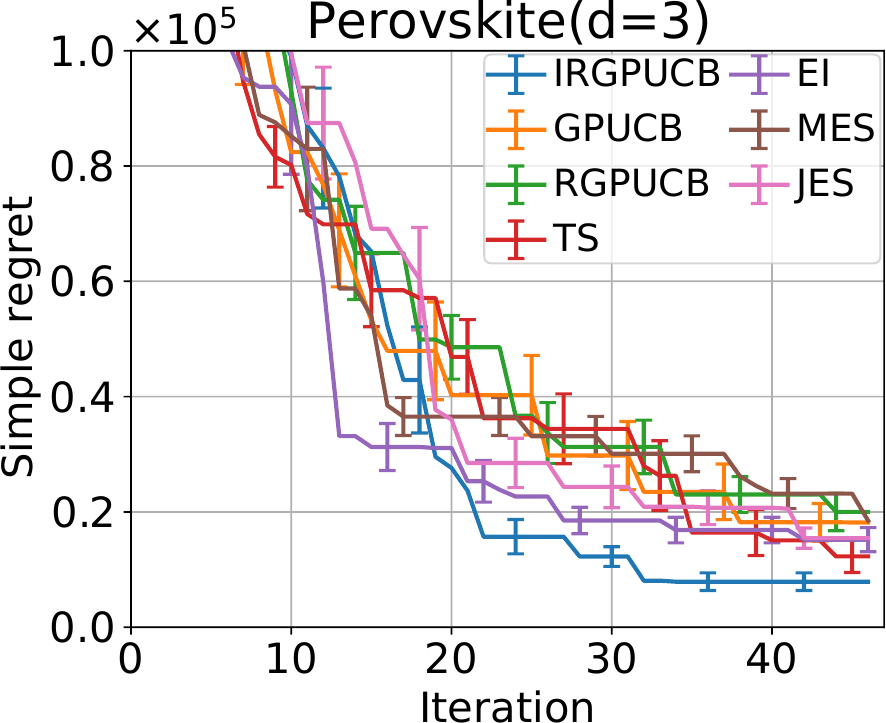}
        \includegraphics[width=0.325\linewidth]{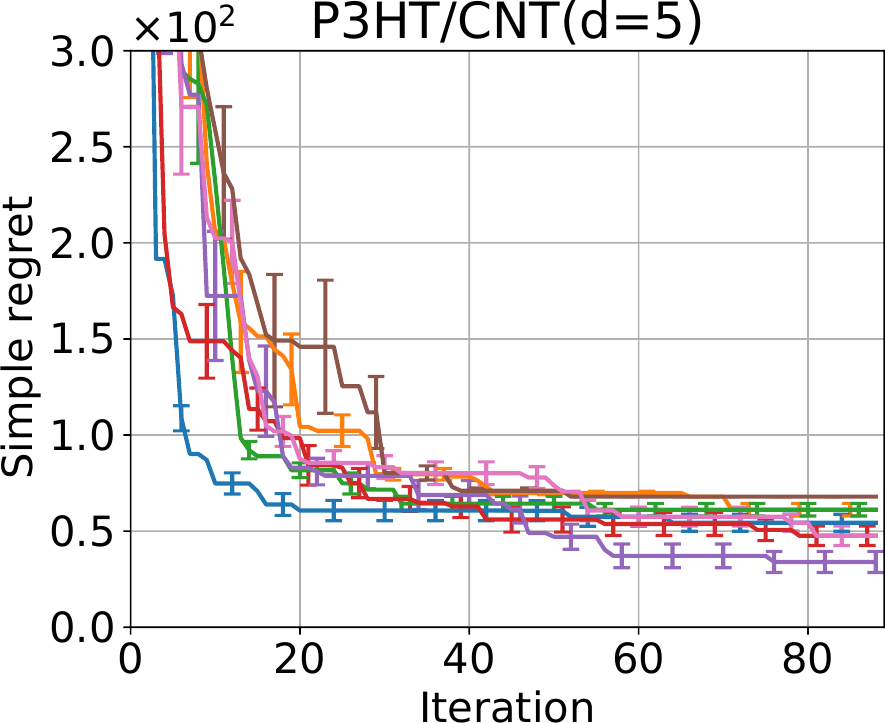}
        \includegraphics[width=0.325\linewidth]{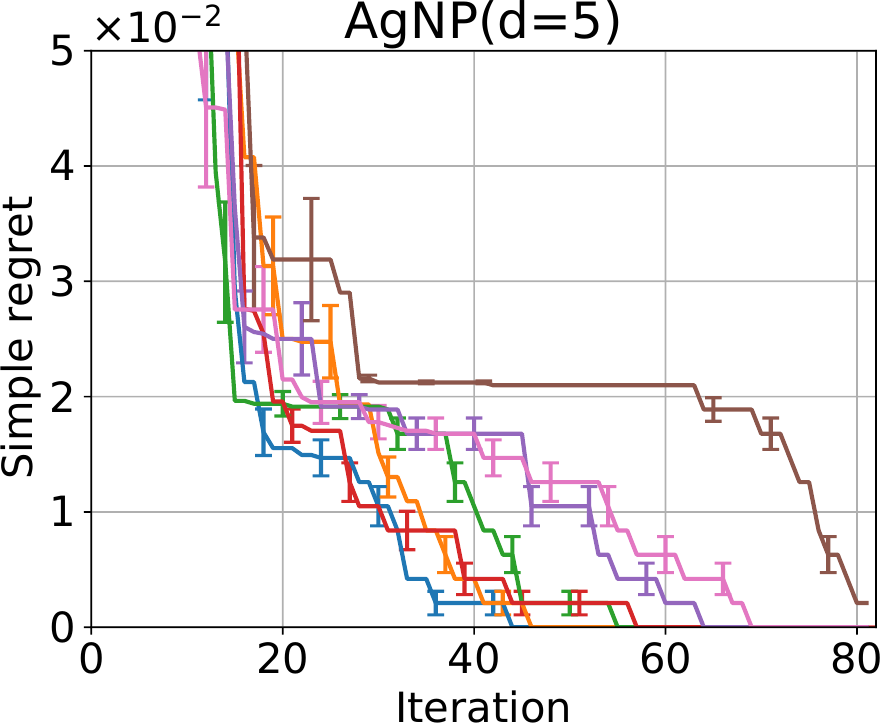}
        \label{fig:real-regret}
        }   
    \vspace{-8pt}
    \caption{Average and standard errors of simple regret.}
\end{figure*}

We employ three benchmark functions called Holder table ($d=2$), Cross in tray($d=2$), and Ackley ($d=4$) functions, whose analytical forms are shown at \url{https://www.sfu.ca/~ssurjano/optimization.html}.
For each function, we report the average and standard error for $10$ trails using ten random initial datasets $\cD_0$, where $|\cD_0| = 2^d$.
We set the noise variance $\sigma^2 = 10^{-4}$.
We used the Gaussian kernel with automatic relevance determination, whose hyperparameter was selected by marginal likelihood maximization per $5$ iterations \citep{Rasmussen2005-Gaussian}.

In this experiment, since the input domain is continuous, the theoretical choice of the confidence parameter contains an unknown variable.
Thus, we use the heuristic choice for confidence parameters.
For GP-UCB, we set the confidence parameter as $\beta_t = 0.2 d \log(2 t)$, which is the heuristics used in \citep{kandasamy2015-high,Kandasamy2017-Multi}.
For RGP-UCB, we set $\zeta_t \sim {\rm Gamma} (\kappa_t, \theta=1)$ with $\kappa_t = 0.2 d \log(2 t)$ since $\EE[\zeta_t]$ must have the same order as $\beta_t$ (note that $\EE[\zeta_t] = \theta \kappa_t$).
For IRGP-UCB, we set $s = d / 2$ and $\lambda = 1 / 2$.
Note that $\lambda$ is equal to the theoretical setting, and $s$ captures the dependence on $d$, as shown in Corollary~\ref{coro:BSR_IRGPUCB_continuous}.

Figure~\ref{fig:benchmark-regret} shows the results.
In the Holder table function, JES shows faster convergence until 40 iterations.
However, the regrets of all baseline methods stagnate in $[10^{-1}, 10^0]$.
In contrast, the regret of IRGP-UCB converges to $10^{-3}$ until 60 iterations.
In the Cross in tray function, IRGP-UCB showed rapid convergence at approximately 50 iterations.
In the Ackley function, IRGP-UCB constantly belonged to the top group and showed minimum regret after 125 iterations.
Since the input dimension is relatively large, TS and JES, which depends on the sampled maxima $\*x_*$, deteriorate by over-exploration.
Throughout the experiments, IRGP-UCB outperforms TS, GP-UCB, and RGP-UCB, which achieves sub-linear BCR bounds, and EI and MES, which are practically well-used.
This result supports the effectiveness of randomization using the two-parameter exponential distribution.

\subsection{Real-World Dataset Experiments}

This section provides the experimental results on the materials datasets provided in \citep{liang2021benchmarking}.
In the perovskite dataset \citep{sun2021data}, we optimize environmental stability with respect to composition parameters for halide perovskite ($d = 3$ and $|\cX| = 94$).
In the P3HT/CNT dataset \citep{bash2021multi}, we optimize electrical conductivity with respect to composition parameters for carbon nanotube polymer blend ($d = 5$ and $|\cX| = 178$).
%
%
In the AgNP dataset \citep{mekki2021two}, we optimize the absorbance spectrum of synthesized silver nanoparticles with respect to processing parameters for synthesizing triangular nanoprisms ($d = 5$ and $|\cX| = 164$).
See \citep{liang2021benchmarking} for more details about each dataset.

We set the initial dataset size $|\cD_0| = 2$ as with \citep{liang2021benchmarking}.
Since the dataset size is small at earlier iterations and the dataset contains fluctuations from real-world experiments, we observed that hyperparameter tuning could be unstable.
Thus, we optimized the hyperparameters of the RBF kernel in each iteration to avoid repeatedly obtaining samples using an inappropriate hyperparameter.
The other settings matched those used in the benchmark function experiments.

Figure~\ref{fig:real-regret} shows the results.
In the perovskite dataset, IRGP-UCB constantly belonged to the top group and showed the best performance after 20 iterations.
In the P3HT/CNT dataset, EI converged to the smallest value after 60 iterations.
On the other hand, IRGP-UCB shows faster convergence during the first 20 iterations.
In the AgNP dataset, IRGP-UCB found the optimal point until 42 iterations in all the trials earliest.
In this experiment, heuristic methods, EI, MES, and JES, showed worse performance and required at least 60 function evaluations to find the optimal point.
Consequently, we can observe the effectiveness of IRGP-UCB against real-world datasets.

\section{Conclusion}
\label{sec:conculusion}

First, by showing the generalized BCR bounds for RGP-UCB, this study showed that randomization without ingenuity does not improve the regret bounds.
Then, this study proposed an improved randomized GP-UCB (IRGP-UCB) using the two-parameter exponential distribution, whose Bayesian regret bound is tighter than known results.
Furthermore, IRGP-UCB does not require an increase in the confidence parameter with respect to the number of iterations.
Lemma~\ref{lem:bound_RGPUCB}, which directly bounds the expectation of the maximum value, plays a key role in the proof.
Additionally, we demonstrated the practical effectiveness of IRGP-UCB through extensive experiments, including the application to the materials dataset.

Several directions for future works can be considered.
First, whether we can show the tighter BCR bounds for the continuous domain is of interest.
Second, since the Bayesian regret analysis of TS depends on the usual UCB, we may improve the BCR bound of TS using randomized UCB.
Last, we may be able to extend IRGP-UCB to the other various practical settings, where the usual UCB-based methods have been extended, e.g., multi-objective BO \citep{paria2020-flexible,Zuluga2016-epsilon,Suzuki2020-multi}, multi-fidelity BO \citep{Kandasamy2016-Gaussian,Kandasamy2017-Multi,Takeno2020-Multifidelity,Takeno2022-generalized}, parallel BO \citep{Contal2013-Parallel,Desautels2014-Parallelizing}, high-dimensional BO \citep{kandasamy2015-high}, cascade BO \citep{Kusakawa2022-bayesian}, and robust BO\citep{Bogunovic2018-adversarially,iwazaki2021-mean,inatsu2022-bayesian}.

\section*{Acknoledements}
This work was supported by MEXT KAKENHI 21H03498, 22H00300, MEXT Program: Data Creation and Utilization-Type Material Research and Development Project Grant Number JPMXP1122712807, and JSPS KAKENHI Grant Number JP20H00601, JP21J14673, and JP23K16943.


\bibliography{ref}

\begin{thebibliography}{43}
\providecommand{\natexlab}[1]{#1}
\providecommand{\url}[1]{\texttt{#1}}
\expandafter\ifx\csname urlstyle\endcsname\relax
  \providecommand{\doi}[1]{doi: #1}\else
  \providecommand{\doi}{doi: \begingroup \urlstyle{rm}\Url}\fi

\bibitem[Bash et~al.(2021)Bash, Cai, Chellappan, Wong, Yang, Kumar, Tan,
  Abutaha, Cheng, Lim, et~al.]{bash2021multi}
D.~Bash, Y.~Cai, V.~Chellappan, S.~L. Wong, X.~Yang, P.~Kumar, J.~D. Tan,
  A.~Abutaha, J.~J. Cheng, Y.-F. Lim, et~al.
\newblock Multi-fidelity high-throughput optimization of electrical
  conductivity in {P3HT-CNT} composites.
\newblock \emph{Advanced Functional Materials}, page 2102606, 2021.

\bibitem[Beg(1980)]{beg1980estimation}
M.~A. Beg.
\newblock On the estimation of pr $\{ Y< X \}$ for the two-parameter
  exponential distribution.
\newblock \emph{Metrika}, 27\penalty0 (1):\penalty0 29--34, 1980.

\bibitem[Berk et~al.(2020)Berk, Gupta, Rana, and
  Venkatesh]{berk2021-randomized}
J.~Berk, S.~Gupta, S.~Rana, and S.~Venkatesh.
\newblock Randomised {G}aussian process upper confidence bound for {B}ayesian
  optimisation.
\newblock In \emph{Proceedings of the 29th International Joint Conference on
  Artificial Intelligence}, pages 2284--2290. International Joint Conferences
  on Artificial Intelligence Organization, 2020.

\bibitem[Bogunovic et~al.(2018)Bogunovic, Scarlett, Jegelka, and
  Cevher]{Bogunovic2018-adversarially}
I.~Bogunovic, J.~Scarlett, S.~Jegelka, and V.~Cevher.
\newblock Adversarially robust optimization with {G}aussian processes.
\newblock In \emph{Advances in Neural Information Processing Systems 31}, pages
  5760--5770. Curran Associates, Inc., 2018.

\bibitem[Bull(2011)]{bull2011convergence}
A.~D. Bull.
\newblock Convergence rates of efficient global optimization algorithms.
\newblock \emph{Journal of Machine Learning Research}, 12\penalty0 (10), 2011.

\bibitem[Chowdhury and Gopalan(2017)]{Chowdhury2017-on}
S.~R. Chowdhury and A.~Gopalan.
\newblock On kernelized multi-armed bandits.
\newblock In \emph{Proceedings of the 34th International Conference on Machine
  Learning}, volume~70 of \emph{Proceedings of Machine Learning Research},
  pages 844--853, 2017.

\bibitem[Contal et~al.(2013)Contal, Buffoni, Robicquet, and
  Vayatis]{Contal2013-Parallel}
E.~Contal, D.~Buffoni, A.~Robicquet, and N.~Vayatis.
\newblock Parallel {G}aussian process optimization with upper confidence bound
  and pure exploration.
\newblock In \emph{Proceedings of the 2013th European Conference on Machine
  Learning and Knowledge Discovery in Databases}, pages 225--240.
  Springer-Verlag, 2013.

\bibitem[Desautels et~al.(2014)Desautels, Krause, and
  Burdick]{Desautels2014-Parallelizing}
T.~Desautels, A.~Krause, and J.~W. Burdick.
\newblock Parallelizing exploration-exploitation tradeoffs in {Gaussian}
  process bandit optimization.
\newblock \emph{Journal of Machine Learning Research}, 15:\penalty0 4053--4103,
  2014.

\bibitem[Griffiths and Hern{\'a}ndez-Lobato(2020)]{griffiths2020-constrained}
R.-R. Griffiths and J.~M. Hern{\'a}ndez-Lobato.
\newblock Constrained {B}ayesian optimization for automatic chemical design
  using variational autoencoders.
\newblock \emph{Chemical science}, 11\penalty0 (2):\penalty0 577--586, 2020.

\bibitem[Hennig and Schuler(2012)]{Henning2012-Entropy}
P.~Hennig and C.~J. Schuler.
\newblock Entropy search for information-efficient global optimization.
\newblock \emph{Journal of Machine Learning Research}, 13\penalty0
  (57):\penalty0 1809--1837, 2012.

\bibitem[Hern\'{a}ndez-Lobato et~al.(2014)Hern\'{a}ndez-Lobato, Hoffman, and
  Ghahramani]{Hernandez2014-Predictive}
J.~M. Hern\'{a}ndez-Lobato, M.~W. Hoffman, and Z.~Ghahramani.
\newblock Predictive entropy search for efficient global optimization of
  black-box functions.
\newblock In \emph{Advances in Neural Information Processing Systems 27}, page
  918–926. Curran Associates, Inc., 2014.

\bibitem[Hvarfner et~al.(2022)Hvarfner, Hutter, and Nardi]{hvarfner2022-joint}
C.~Hvarfner, F.~Hutter, and L.~Nardi.
\newblock Joint entropy search for maximally-informed {B}ayesian optimization.
\newblock In \emph{Advances in Neural Information Processing Systems}, 2022.
\newblock To appear.

\bibitem[Inatsu et~al.(2022)Inatsu, Takeno, Karasuyama, and
  Takeuchi]{inatsu2022-bayesian}
Y.~Inatsu, S.~Takeno, M.~Karasuyama, and I.~Takeuchi.
\newblock {B}ayesian optimization for distributionally robust
  chance-constrained problem.
\newblock In \emph{Proceedings of the 39th International Conference on Machine
  Learning}, volume 162, pages 9602--9621. PMLR, 2022.

\bibitem[Iwazaki et~al.(2021)Iwazaki, Inatsu, and Takeuchi]{iwazaki2021-mean}
S.~Iwazaki, Y.~Inatsu, and I.~Takeuchi.
\newblock Mean-variance analysis in {B}ayesian optimization under uncertainty.
\newblock In \emph{Proceedings of The 24th International Conference on
  Artificial Intelligence and Statistics}, volume 130, pages 973--981. PMLR,
  2021.

\bibitem[Janz et~al.(2020)Janz, Burt, and Gonzalez]{janz2020-bandit}
D.~Janz, D.~Burt, and J.~Gonzalez.
\newblock Bandit optimisation of functions in the {Matérn kernel RKHS}.
\newblock In \emph{Proceedings of the 23rd International Conference on
  Artificial Intelligence and Statistics}, volume 108 of \emph{Proceedings of
  Machine Learning Research}, pages 2486--2495, 2020.

\bibitem[Kandasamy et~al.(2015)Kandasamy, Schneider, and
  Poczos]{kandasamy2015-high}
K.~Kandasamy, J.~Schneider, and B.~Poczos.
\newblock High dimensional {B}ayesian optimisation and bandits via additive
  models.
\newblock In \emph{Proceedings of the 32nd International Conference on Machine
  Learning}, volume~37 of \emph{Proceedings of Machine Learning Research},
  pages 295--304, 2015.

\bibitem[Kandasamy et~al.(2016)Kandasamy, Dasarathy, Oliva, Schneider, and
  P{\'o}czos]{Kandasamy2016-Gaussian}
K.~Kandasamy, G.~Dasarathy, J.~Oliva, J.~Schneider, and B.~P{\'o}czos.
\newblock Gaussian process bandit optimisation with multi-fidelity evaluations.
\newblock In \emph{Advances in Neural Information Processing Systems 29}, pages
  1000--1008. Curran Associates, Inc., 2016.

\bibitem[Kandasamy et~al.(2017)Kandasamy, Dasarathy, Schneider, and
  P{\'o}czos]{Kandasamy2017-Multi}
K.~Kandasamy, G.~Dasarathy, J.~Schneider, and B.~P{\'o}czos.
\newblock Multi-fidelity {B}ayesian optimisation with continuous
  approximations.
\newblock In \emph{Proceedings of the 34th International Conference on Machine
  Learning}, volume~70 of \emph{Proceedings of Machine Learning Research},
  pages 1799--1808, 2017.

\bibitem[Kandasamy et~al.(2018{\natexlab{a}})Kandasamy, Krishnamurthy,
  Schneider, and P{\'o}czos]{Kandasamy2018-Parallelised}
K.~Kandasamy, A.~Krishnamurthy, J.~Schneider, and B.~P{\'o}czos.
\newblock Parallelised {B}ayesian optimisation via {T}hompson sampling.
\newblock In \emph{Proceedings of the 21st International Conference on
  Artificial Intelligence and Statistics}, volume~84 of \emph{Proceedings of
  Machine Learning Research}, pages 133--142, 2018{\natexlab{a}}.

\bibitem[Kandasamy et~al.(2018{\natexlab{b}})Kandasamy, Neiswanger, Schneider,
  Poczos, and Xing]{kandasamy2018-neural}
K.~Kandasamy, W.~Neiswanger, J.~Schneider, B.~Poczos, and E.~P. Xing.
\newblock Neural architecture search with {B}ayesian optimisation and optimal
  transport.
\newblock \emph{Advances in neural information processing systems 31}, pages
  2016--2025, 2018{\natexlab{b}}.

\bibitem[Korovina et~al.(2020)Korovina, Xu, Kandasamy, Neiswanger, Poczos,
  Schneider, and Xing]{korovina2020-chembo}
K.~Korovina, S.~Xu, K.~Kandasamy, W.~Neiswanger, B.~Poczos, J.~Schneider, and
  E.~Xing.
\newblock {ChemBO: B}ayesian optimization of small organic molecules with
  synthesizable recommendations.
\newblock In \emph{Proceedings of the 23rd International Conference on
  Artificial Intelligence and Statistics}, volume 108 of \emph{Proceedings of
  Machine Learning Research}, pages 3393--3403, 2020.

\bibitem[Kusakawa et~al.(2022)Kusakawa, Takeno, Inatsu, Kutsukake, Iwazaki,
  Nakano, Ujihara, Karasuyama, and Takeuchi]{Kusakawa2022-bayesian}
S.~Kusakawa, S.~Takeno, Y.~Inatsu, K.~Kutsukake, S.~Iwazaki, T.~Nakano,
  T.~Ujihara, M.~Karasuyama, and I.~Takeuchi.
\newblock {B}ayesian optimization for cascade-type multistage processes.
\newblock \emph{Neural Computation}, 34\penalty0 (12):\penalty0 2408--2431,
  2022.

\bibitem[Lam et~al.(1994)Lam, Sinha, and Wu]{lam1994estimation}
K.~Lam, B.~K. Sinha, and Z.~Wu.
\newblock Estimation of parameters in a two-parameter exponential distribution
  using ranked set sample.
\newblock \emph{Annals of the Institute of Statistical Mathematics},
  46\penalty0 (4):\penalty0 723--736, 1994.

\bibitem[Liang et~al.(2021)Liang, Gongora, Ren, Tiihonen, Liu, Sun, Deneault,
  Bash, Mekki-Berrada, Khan, et~al.]{liang2021benchmarking}
Q.~Liang, A.~Gongora, Z.~Ren, A.~Tiihonen, Z.~Liu, S.~Sun, J.~Deneault,
  D.~Bash, F.~Mekki-Berrada, S.~Khan, et~al.
\newblock Benchmarking the performance of {B}ayesian optimization across
  multiple experimental materials science domains.
\newblock \emph{npj Computational Material}, 7\penalty0 (188), 2021.

\bibitem[Mekki-Berrada et~al.(2021)Mekki-Berrada, Ren, Huang, Wong, Zheng, Xie,
  Tian, Jayavelu, Mahfoud, Bash, et~al.]{mekki2021two}
F.~Mekki-Berrada, Z.~Ren, T.~Huang, W.~K. Wong, F.~Zheng, J.~Xie, I.~P.~S.
  Tian, S.~Jayavelu, Z.~Mahfoud, D.~Bash, et~al.
\newblock Two-step machine learning enables optimized nanoparticle synthesis.
\newblock \emph{npj Computational Materials}, 7\penalty0 (1):\penalty0 1--10,
  2021.

\bibitem[Mockus et~al.(1978)Mockus, Tiesis, and
  Zilinskas]{Mockus1978-Application}
J.~Mockus, V.~Tiesis, and A.~Zilinskas.
\newblock The application of {B}ayesian methods for seeking the extremum.
\newblock \emph{Towards Global Optimization}, 2\penalty0 (117-129):\penalty0 2,
  1978.

\bibitem[Paria et~al.(2020)Paria, Kandasamy, and
  P{\'{o}}czos]{paria2020-flexible}
B.~Paria, K.~Kandasamy, and B.~P{\'{o}}czos.
\newblock A flexible framework for multi-objective {B}ayesian optimization
  using random scalarizations.
\newblock In \emph{Proceedings of The 35th Uncertainty in Artificial
  Intelligence Conference}, volume 115 of \emph{Proceedings of Machine Learning
  Research}, pages 766--776, 2020.

\bibitem[Rahimi and Recht(2008)]{Rahimi2008-Random}
A.~Rahimi and B.~Recht.
\newblock Random features for large-scale kernel machines.
\newblock In \emph{Advances in Neural Information Processing Systems 20}, pages
  1177--1184. Curran Associates, Inc., 2008.

\bibitem[Rasmussen and Williams(2005)]{Rasmussen2005-Gaussian}
C.~E. Rasmussen and C.~K.~I. Williams.
\newblock \emph{Gaussian Processes for Machine Learning (Adaptive Computation
  and Machine Learning)}.
\newblock The MIT Press, 2005.

\bibitem[Russo and Van~Roy(2014)]{Russo2014-learning}
D.~Russo and B.~Van~Roy.
\newblock Learning to optimize via posterior sampling.
\newblock \emph{Mathematics of Operations Research}, 39\penalty0 (4):\penalty0
  1221--1243, 2014.

\bibitem[Shahriari et~al.(2016)Shahriari, Swersky, Wang, Adams, and {De
  Freitas}]{Shahriari2016-Taking}
B.~Shahriari, K.~Swersky, Z.~Wang, R.~Adams, and N.~{De Freitas}.
\newblock Taking the human out of the loop: A review of {B}ayesian
  optimization.
\newblock \emph{Proceedings of the IEEE}, 104\penalty0 (1):\penalty0 148--175,
  2016.

\bibitem[Snoek et~al.(2012)Snoek, Larochelle, and Adams]{Snoek2012-Practical}
J.~Snoek, H.~Larochelle, and R.~P. Adams.
\newblock Practical {B}ayesian optimization of machine learning algorithms.
\newblock In \emph{Advances in Neural Information Processing Systems 25}, pages
  2951--2959. Curran Associates, Inc., 2012.

\bibitem[Srinivas et~al.(2010)Srinivas, Krause, Kakade, and
  Seeger]{Srinivas2010-Gaussian}
N.~Srinivas, A.~Krause, S.~Kakade, and M.~Seeger.
\newblock Gaussian process optimization in the bandit setting: No regret and
  experimental design.
\newblock In \emph{Proceedings of the 27th International Conference on Machine
  Learning}, pages 1015--1022. Omnipress, 2010.

\bibitem[Sun et~al.(2021)Sun, Tiihonen, Oviedo, Liu, Thapa, Zhao, Hartono,
  Goyal, Heumueller, Batali, et~al.]{sun2021data}
S.~Sun, A.~Tiihonen, F.~Oviedo, Z.~Liu, J.~Thapa, Y.~Zhao, N.~T.~P. Hartono,
  A.~Goyal, T.~Heumueller, C.~Batali, et~al.
\newblock A data fusion approach to optimize compositional stability of halide
  perovskites.
\newblock \emph{Matter}, 4\penalty0 (4):\penalty0 1305--1322, 2021.

\bibitem[Suzuki et~al.(2020)Suzuki, Takeno, Tamura, Shitara, and
  Karasuyama]{Suzuki2020-multi}
S.~Suzuki, S.~Takeno, T.~Tamura, K.~Shitara, and M.~Karasuyama.
\newblock Multi-objective {B}ayesian optimization using {P}areto-frontier
  entropy.
\newblock In \emph{Proceedings of the 37th International Conference on Machine
  Learning}, volume 119, pages 9279--9288. PMLR, 2020.

\bibitem[Takeno et~al.(2020)Takeno, Fukuoka, Tsukada, Koyama, Shiga, Takeuchi,
  and Karasuyama]{Takeno2020-Multifidelity}
S.~Takeno, H.~Fukuoka, Y.~Tsukada, T.~Koyama, M.~Shiga, I.~Takeuchi, and
  M.~Karasuyama.
\newblock Multi-fidelity {B}ayesian optimization with max-value entropy search
  and its parallelization.
\newblock In \emph{Proceedings of the 37th International Conference on Machine
  Learning}, volume 119, pages 9334--9345. PMLR, 2020.

\bibitem[Takeno et~al.(2022{\natexlab{a}})Takeno, Fukuoka, Tsukada, Koyama,
  Shiga, Takeuchi, and Karasuyama]{Takeno2022-generalized}
S.~Takeno, H.~Fukuoka, Y.~Tsukada, T.~Koyama, M.~Shiga, I.~Takeuchi, and
  M.~Karasuyama.
\newblock {A Generalized Framework of Multifidelity Max-Value Entropy Search
  Through Joint Entropy}.
\newblock \emph{Neural Computation}, 34\penalty0 (10):\penalty0 2145--2203,
  2022{\natexlab{a}}.

\bibitem[Takeno et~al.(2022{\natexlab{b}})Takeno, Tamura, Shitara, and
  Karasuyama]{takeno2022-sequential}
S.~Takeno, T.~Tamura, K.~Shitara, and M.~Karasuyama.
\newblock Sequential and parallel constrained max-value entropy search via
  information lower bound.
\newblock In \emph{Proceedings of the 39th International Conference on Machine
  Learning}, volume 162 of \emph{Proceedings of Machine Learning Research},
  pages 20960--20986, 2022{\natexlab{b}}.

\bibitem[Ueno et~al.(2016)Ueno, Rhone, Hou, Mizoguchi, and
  Tsuda]{ueno2016combo}
T.~Ueno, T.~D. Rhone, Z.~Hou, T.~Mizoguchi, and K.~Tsuda.
\newblock {COMBO}: An efficient {B}ayesian optimization library for materials
  science.
\newblock \emph{Materials discovery}, 4:\penalty0 18--21, 2016.

\bibitem[Vakili et~al.(2021)Vakili, Khezeli, and
  Picheny]{vakili2021-information}
S.~Vakili, K.~Khezeli, and V.~Picheny.
\newblock On information gain and regret bounds in {G}aussian process bandits.
\newblock In \emph{Proceedings of The 24th International Conference on
  Artificial Intelligence and Statistics}, volume 130 of \emph{Proceedings of
  Machine Learning Research}, pages 82--90, 2021.

\bibitem[Wang and Jegelka(2017)]{Wang2017-Max}
Z.~Wang and S.~Jegelka.
\newblock Max-value entropy search for efficient {B}ayesian optimization.
\newblock In \emph{Proceedings of the 34th International Conference on Machine
  Learning}, volume~70 of \emph{Proceedings of Machine Learning Research},
  pages 3627--3635, 2017.

\bibitem[Wang et~al.(2016)Wang, Zhou, and Jegelka]{Wang2016-Optimization}
Z.~Wang, B.~Zhou, and S.~Jegelka.
\newblock Optimization as estimation with {G}aussian processes in bandit
  settings.
\newblock In \emph{Proceedings of the 19th International Conference on
  Artificial Intelligence and Statistics}, volume~51 of \emph{Proceedings of
  Machine Learning Research}, pages 1022--1031, 2016.

\bibitem[Zuluaga et~al.(2016)Zuluaga, Krause, and
  P{{{\"u}}}schel]{Zuluga2016-epsilon}
M.~Zuluaga, A.~Krause, and M.~P{{{\"u}}}schel.
\newblock {e-PAL}: An active learning approach to the multi-objective
  optimization problem.
\newblock \emph{Journal of Machine Learning Research}, 17\penalty0
  (104):\penalty0 1--32, 2016.

\end{thebibliography}
\bibliographystyle{abbrvnat}
\clearpage
\appendix
\onecolumn

\section{BSR with Recommended Input}
\label{app:BSR_recommendation}
When the observation is contaminated by the noise, we cannot know the observed input, which minimizes BSR defined in Eq.~\eqref{eq:BSR}.
Thus, it is required that the algorithm explicitly returns low regret input.
Then, we can consider the following variant of BSR:
\begin{align*}
    {\rm BSR}_T \coloneqq \EE\bigl[ f(\*x^*) - f(\hat{\*x}_T) \bigr],
\end{align*}
where $\hat{\*x}_t$ is a recommendation point from the algorithm.
To bound this BSR, we can consider the following two options for the recommendation:
\begin{align*}
    \hat{\*x}_t &= \argmax_{\*x \in \cX} \mu_{t-1} (\*x), \\
    \hat{\*x}_t &=\argmax_{\*x \in \{ \*x_1, \dots, \*x_{t-1} \} } \mu_{t-1} (\*x).
\end{align*}
Although we here consider the finite input domain for simplicity, it can be extended to the infinite domain in the same way as our proof of Corollaries~\ref{coro:BSR_IRGPUCB_continuous}.
Then, the BSR defined by the both $\hat{\*x}_t$ can be bounded as follows:
\begin{align*}
    {\rm BSR}_T 
    &= \EE\bigl[ f(\*x^*) - f(\hat{\*x}_T) \bigr] \\
    &= \EE_{\cD_{T-1}} \bigl[ \EE_f \bigl[ f(\*x^*) | \cD_{T-1} \bigr] - \mu_{T-1} (\hat{\*x}_T) \bigr] \\
    &\leq \EE_{\cD_{T-1}} \biggl[ \EE_f \bigl[ f(\*x^*) | \cD_{T-1} \bigr] - \frac{1}{T} \sum_{t=1}^{T} \mu_{T-1} (\*x_t) \biggr] && \bigl(\because \forall 1 \leq t \leq T, \mu_{T-1} (\hat{\*x}_T) \leq \mu_{T-1} (\*x_t)  \bigr) \\
    &= \frac{1}{T} \sum_{t=1}^{T} \EE_{\cD_{T-1}} \Bigl[ \EE_f \bigl[ f(\*x^*) | \cD_{T-1} \bigr] - \EE_f \bigl[ f(\*x_t) | \cD_{T-1} \bigr] \Bigr] \\
    &\leq \frac{1}{T} \sum_{t=1}^{T} \EE \bigl[ f(\*x^*) -   f (\*x_t) \bigr] \\
    &= \frac{1}{T} {\rm BCR}_T.
\end{align*}
Consequently, we can show the upper bound of this variant of BSR, which is the same as that of the BSR defined in Eq.~\eqref{eq:BSR}.

\section{Proof for BCR Bound of GP-UCB}
\label{app:GP_UCB_proof}

\citet{Srinivas2010-Gaussian} provided the high-probability bounds for cumulative regret, not BCR bounds.
On the other hand, although \citet{paria2020-flexible} have shown BCR bounds for the GP-UCB-based multi-objective BO method, they did not provide explicit proof for standard single-objective BO.
Thus, although the following result is an almost direct consequence of Theorem~2 in \citep{paria2020-flexible}, we here provide the explicit proof for standard GP-UCB for completeness.
On the other hand, for the case of (ii), the bound with respect to $a$ is tightened as $O\bigl( \log ( \log (a)) \bigr)$ compared with $O( \log(a))$ \citep{Kandasamy2018-Parallelised, paria2020-flexible}, which is mainly based on Lemma~\ref{lem:L_max_bound}.

\begin{theorem}
    Suppose that $f \sim \cG \cP(0, k)$, where $k$ is a stationary kernel and $k(\*x, \*x) = 1$.
    
    \vspace{-0.8\baselineskip}
    \begin{itemize}
        \setlength{\parskip}{0.05cm} 
        \setlength{\itemsep}{0.05cm} 
        \item[(i)] When $\cX$ is a finite set, GP-UCB with $\beta_t = 2 \log(|\cX| t^2 / \sqrt{2\pi} )$ achieves the following BCR bound:
        \begin{align}
            {\rm BCR}_T \leq \frac{\pi^2}{6} + \sqrt{ C_1 T \beta_T \gamma_T},
        \end{align}
        where $C_1 \coloneqq 2 / \log(1 + \sigma^{-2})$. 
        \item[(ii)] When $\cX$ is an infinite set, let Assumption~\ref{assump:continuous_X} holds. 
                    Then, GP-UCB with $\beta_t = 2d \log \bigl(bdr t^2 \bigl( \sqrt{\log (ad)} + \sqrt{\pi} / 2 \bigr) \bigr) + 2\log (t^2 / \sqrt{2\pi})$ achieves the following BCR bound:
        \begin{align}
            {\rm BCR}_T \leq \frac{\pi^2}{3} + \sqrt{ C_1 T \beta_T \gamma_T},
        \end{align}
        where $C_1 \coloneqq 2 / \log(1 + \sigma^{-2})$. 
    \end{itemize}
    \vspace{-0.8\baselineskip}
    \label{theo:BCR_GPUCB}
\end{theorem}
\begin{proof}
    First, we show the proof of (i).
    For regret analysis of BCR, regret decomposition  \citep{Russo2014-learning} with $U_t(\*x) \coloneqq \mu_{t-1}(\*x) + \beta_t^{1/2} \sigma_{t-1}(\*x)$ is used as follows:
    \begin{align}
        {\rm BCR}_T 
        &= \sum_{t=1}^T \EE [f(\*x^*) - f(\*x_t)] \\
        &= \sum_{t=1}^T \EE [f(\*x^*) - U_t(\*x^*) + U_t(\*x^*) - U_t(\*x_t) + U_t(\*x_t) - f(\*x_t)] \\
        &\leq \underbrace{\sum_{t=1}^T \EE [f(\*x^*) - U_t(\*x^*)]}_{R_1} + \underbrace{\sum_{t=1}^T \EE [ U_t(\*x_t) - f(\*x_t)]}_{R_2}. && \bigl( \because U_t(\*x^*) - U_t(\*x_t) \leq 0 \bigr)
    \end{align}

    First, we consider $R_1$.
    We use the following fact for a Gaussian random variable $Z \sim \cN(m, s^2)$ with $m \leq 0$:
    \begin{align}
        \EE[Z_+] \leq \frac{s}{\sqrt{2 \pi}} \exp \left\{ - \frac{m^2}{2 s^2} \right\}, \label{eq:Gauss_plus_expectation_1}
    \end{align}
    where $Z_+ \coloneqq \max \{0, Z\}$.
    This fact has been frequently used in this literature \citep{Russo2014-learning,Kandasamy2018-Parallelised,paria2020-flexible}.
    Then, we can extend $R_1$ as follows:
    \begin{align}
        R_1 
        &= \sum_{t=1}^T \EE_{\cD_{t-1}} [ \EE_t[ f(\*x^*) - U_t(\*x^*)]] \\
        &\leq \sum_{t=1}^T \EE_{\cD_{t-1}} [ \EE_t[ \bigl( f(\*x^*) - U_t(\*x^*) \bigr)_+ ]] && \bigl( \because f(\*x^*) - U_t(\*x^*) \leq \bigl( f(\*x^*) - U_t(\*x^*) \bigr)_+ \bigr) \\
        &\leq \sum_{t=1}^T \sum_{\*x \in \cX} \EE_{\cD_{t-1}} \left[ \EE_t \left[ \bigl(f(\*x) - U_t(\*x)\bigr)_+  \right] \right]. && \left( \because \bigl( f(\*x^*) - U_t(\*x^*) \bigr)_+ \leq \sum_{\*x \in \cX} \bigl( f(\*x) - U_t(\*x) \bigr)_+ \right)
    \end{align}
    Then, since $f(\*x) - U_t(\*x) \mid \cD_{t-1} \sim \cN( -\beta_t^{1/2}(\*x)\sigma_{t-1}(\*x), \sigma^2_{t-1}(\*x))$, we can apply Eq.~\eqref{eq:Gauss_plus_expectation_1}:
    \begin{align}
        R_1
        &\leq \sum_{t=1}^T \sum_{\*x \in \cX} \EE_{\cD_{t-1}} \left[ \frac{\sigma_{t-1}(\*x)}{\sqrt{2 \pi}} \exp \left\{ - \frac{\beta_t}{2} \right\} \right] && \bigl( \because \text{Eq.~\eqref{eq:Gauss_plus_expectation_1}} \bigr) \\
        &\leq \sum_{t=1}^T |\cX| \frac{1}{\sqrt{2 \pi}} \exp \left\{ - \frac{\beta_t}{2} \right\} && \bigl( \because \sigma_{t-1}(\*x) \leq \sigma_0(\*x) = 1 \bigr) \\
        &= \sum_{t=1}^T \frac{1}{t^2} \\
        &\leq \frac{\pi^2}{6}. && \left(\because \sum_{t=1}^\infty \frac{1}{t^2} = \frac{\pi^2}{6} \right) \label{eq:GP_UCB_R1_bound} 
    \end{align}

    Then, we bound $R_2$ as follows:
    \begin{align}
        R_2
        &= \sum_{t=1}^T \EE [ U_t(\*x_t) - f(\*x_t)] \\
        &= \sum_{t=1}^T \EE_{\cD_{t-1}} [ \EE_t [ U_t(\*x_t) - f(\*x_t)] ] \\
        &= \sum_{t=1}^T \EE_{\cD_{t-1}} [ U_t(\*x_t) - \mu_{t-1}(\*x_t) ] \\
        &= \sum_{t=1}^T \EE \left[ \beta^{1/2}_t \sigma_{t-1}(\*x_t) \right] \\
        &= \EE \left[ \sum_{t=1}^T \beta^{1/2}_t \sigma_{t-1}(\*x_t) \right] \\
        &\leq \EE \left[ \sqrt{ \sum_{t=1}^T \beta_t  \sum_{t=1}^T \sigma^2_{t-1}(\*x_t) } \right] && \bigl( \because \text{Cauchy--Schwarz  inequality} \bigr) \\
        &= \sqrt{ \sum_{t=1}^T \beta_t} \EE \left[ \sqrt{ \sum_{t=1}^T \sigma^2_{t-1}(\*x_t) } \right] \\
        &\leq \sqrt{ T \beta_T} \EE \left[ \sqrt{ \sum_{t=1}^T \sigma^2_{t-1}(\*x_t) } \right] && \bigl( \because \beta_1 < \beta_2 < \dots < \beta_T \bigr).
    \end{align}
    Then, from Lemma~5.4 in \citep{Srinivas2010-Gaussian}, $\sqrt{\sum_{t=1}^T \sigma^2_{t-1}(\*x_t)} \leq \sqrt{C_1 \gamma_T}$ with probability $1$.
    Thus, we obtain
    \begin{align}
        R_2 \leq \sqrt{ C_1 T \beta_T \gamma_T}. \label{eq:GP_UCB_R2_bound}
    \end{align}
    Consequently, by combining Eqs.~\eqref{eq:GP_UCB_R1_bound} and \eqref{eq:GP_UCB_R2_bound}, we obtain the theorem for (i).

    Next, we consider (ii).
    Purely for the sake of analysis, we use a set of discretization $\cX_t \subset \cX$ for $t \geq 1$.
    For any $t \geq 1$, let $\cX_t \subset \cX$ be a finite set with each dimension equally divided into $\tau_t = bdr t^2 \bigl(\sqrt{\log (ad)} + \sqrt{\pi} / 2 \bigr)$.
    Thus, $|\cX_t| = \tau_t^d$.
    In addition, we define $[\*x]_t$ as the nearest points in $\cX_t$ of $\*x \in \cX$.

    Then, BCR can be decomposed as follows:
    \begin{align*}
        {\rm BCR}_T 
        &= \sum_{t=1}^T \EE \left[f(\*x^*) - f([\*x^*]_t) + f([\*x^*]_t) - U_t([\*x^*]_t) + U_t([\*x^*]_t) - U_t(\*x_t) + U_t(\*x_t) - f(\*x_t) \right] \\
        &\leq \sum_{t=1}^T \EE \left[f(\*x^*) - f([\*x^*]_t) + f([\*x^*]_t) - U_t([\*x^*]_t) + U_t(\*x_t) - f(\*x_t) \right],
    \end{align*}
    where $\EE \bigl[ U_t([\*x^*]_t) - U_t(\*x_t) \bigr] \leq 0$ since $\*x_t = \argmax_{\*x \in \cX} U_t (\*x)$.
    Furthermore, we obtain the following:
    \begin{align*}
        \sum_{t=1}^T \EE \left[f(\*x^*) - f([\*x^*]_t) \right]
        &\leq \sum_{t=1}^T \EE \left[ \sup_{\*x \in \cX} f(\*x) - f([\*x]_t) \right] \\
        &\leq \sum_{t=1}^T \frac{1}{t^2} && \left( \because \text{Lemma}~\ref{lem:discretized_error} \right) \\
        &\leq \frac{\pi^2}{6} && \left(\because \sum_{t=1}^\infty \frac{1}{t^2} = \frac{\pi^2}{6} \right)
    \end{align*}
    The remaining terms $\sum_{t=1}^T \EE\bigl[ f([\*x^*]_t) - U_t([\*x^*]_t) + U_t(\*x_t) - f(\*x_t) \bigr]$ can be bounded as with the case (i) by setting $\beta_t = 2 \log \bigl( |\cX_t| t^2 / \sqrt{2\pi} \bigr)$.
    That is, 
    \begin{align}
        \sum_{t=1}^T \EE\bigl[ f([\*x^*]_t) - U_t([\*x^*]_t) + U_t(\*x_t) - f(\*x_t) \bigr]
        &\leq \frac{\pi^2}{6} + \sqrt{C_1 T \beta_T \gamma_T}.
    \end{align}
    By substituting $|\cX_t| = \bigl(  bdr t^2 \bigl(\sqrt{\log (ad)} + \sqrt{\pi} / 2 \bigr)\bigr)^d$, we conclude the proof.
\end{proof}

\section{Technical Issues of Proof in \citep{berk2021-randomized}}
\label{app:issues_RGPUCB}

The proofs of Theorems~1, 2, and 3 in \citep{berk2021-randomized} appear to contain many technical issues.
We enumerated them using the notation in \citep{berk2021-randomized} as follows:
\begin{itemize}
    \item 
        Lemma~1 does not necessarily hold since $\alpha([\*x_t]_\tau)$ can be smaller than $\alpha([\*x^*]_\tau)$ although $\alpha(\*x_t) \geq \alpha(\*x^*)$.
    \item 
        To treat $f(\*x_t)$ as the random variable, which follows $\cN(\mu_t(\*x), \sigma^2_t(\*x))$, the conditioning by $D_t$ is required.
        However, they do not perform the conditioning in most equations, such as Eqs.~(11) and (12). 
        %
    %
    \item 
        From Eq.~(13) to the next equations, the term $|\cX_{\rm dis}|$ is eliminated. 
        This term is imperatively required.
    \item 
        In Eqs.~(11), (12), (16), and (17), variables such as $\sigma_{t-1}([\*x_t]_r)$ are random variables that depend on $D_t$ and $[\*x_t]_r$. 
        However, these random variables are taken out from the expectation.
    \item 
        From Eq.(17) to Eq.~(19), a discretization for $\*x_t$ is eliminated.
    \item 
        In Eq.~(18), an approximately equal sign $\approx$ is used, which is inappropriate for theoretical proof.
    \item 
        The convergence rate of the term $F^{-1}(1 - 1/T)$ in the resulting bound is not clarified.
\end{itemize}
Hence, Theorems~1, 2, and 3 in \citep{berk2021-randomized} are not shown using their proofs.
Furthermore, even if Theorem~3 holds, it would be insufficient to show the sub-linear BCR bound.

\section{Proof of Theorem for RGP-UCB}
\label{app:proof_general_RGPUCB}

Here, we rectify and generalize the theorems in \citep{berk2021-randomized}.
First, we show the generalized theorem, which shows the upper bound using the expectation and MGFs of $\{ \zeta_t \}_{t \geq 1}$.
Next, we provide some examples of the distributions, which can achieve the sub-linear regret bound by appropriately controlling the parameters.

\subsection{Generalized BCR Bounds for RGP-UCB}

The following theorem shows the upper bound of BCR for setting $\{\zeta_t\}_{t \geq 1}$ as arbitrary non-negative random variables:
\begin{theorem}
    Let $f \sim \cG \cP(0, k)$, where $k$ is a stationary kernel $k$ and $k(\*x, \*x) = 1$.
    Let $\{\zeta_t\}_{t \geq 1}$ be a sequence of non-negative random variables and their MGFs be $M_t (\cdot)$, where $M_t (-1/2) < \infty$ at least.
    \vspace{-0.8\baselineskip}
    \begin{itemize}
        \setlength{\parskip}{0.05cm} 
        \setlength{\itemsep}{0.05cm} 
        \item[(i)] When $\cX$ is a finite set, by running RGP-UCB with $\zeta_t$, BCR can be bounded as follows: 
            \begin{align*}
                {\rm BCR}_T \leq \frac{ |\cX|}{\sqrt{2 \pi}} \sum_{t=1}^T M_t(-1/2) + \sqrt{ \sum_{t=1}^T \EE[\zeta_t]  C_1 \gamma_T},
            \end{align*}
            where $C_1 \coloneqq 2 / \log(1 + \sigma^{-2})$. 
        \item[(ii)] When $\cX$ is an infinite set, let Assumption~\ref{assump:continuous_X} holds. 
                    By running RGP-UCB with $\zeta_t$, BCR can be bounded as follows: 
            \begin{align}
                {\rm BCR}_T \leq \frac{\pi^2}{6} +  \frac{1}{\sqrt{2 \pi}} \sum_{t=1}^T (\tau_t)^d M_t(-1/2)  + \sqrt{ C_1 \sum_{t=1}^T \EE[\zeta_t] \gamma_T},
            \end{align}
            where $C_1 \coloneqq 2 / \log(1 + \sigma^{-2})$ and $\tau_t = bdr t^2 \bigl(\sqrt{\log (ad)} + \sqrt{\pi} / 2 \bigr)$. 
    \end{itemize}
    \vspace{-0.8\baselineskip}
    \label{theo:BCR_RGPUCB}
\end{theorem}
\begin{proof}
    The procedure of the proof is similar to the proof for the standard GP-UCB shown in Appendix~\ref{app:GP_UCB_proof}.
    BCR can be decomposed by $V_t(\*x) = \mu_{t-1}(\*x) + \zeta_t^{1/2} \sigma_{t-1}(\*x)$ as follows:
    \begin{align*}
        {\rm BCR}_T 
        &= \sum_{t=1}^T \EE \left[ f(\*x^*) - V_t(\*x^*) + V_t(\*x^*) - V_t(\*x_t) + V_t(\*x_t) - f(\*x_t) \right] \\
        &\leq \underbrace{\sum_{t=1}^T \EE \left[ f(\*x^*) - V_t(\*x^*) \right]}_{\eqqcolon R_1} + \underbrace{\sum_{t=1}^T \EE\left[ V_t(\*x_t) - f(\*x_t) \right]}_{ \eqqcolon R_2}. && \bigl(\because V_t(\*x^*) - V_t(\*x_t) \leq 0 \bigr)
    \end{align*}
    Then, we derive the upper bounds of $R_1$ and $R_2$, respectively.

    First, we consider $R_1$.
    We use the following fact for Gaussian random variables $Z \sim \cN(m, s^2)$ with $m \leq 0$:
    \begin{align}
        \EE[Z_+] \leq \frac{s}{\sqrt{2 \pi}} \exp \left\{ - \frac{m^2}{2 s^2} \right\}, \label{eq:Gauss_plus_expectation}
    \end{align}
    where $Z_+ \coloneqq \max \{0, Z\}$.
    This fact is frequently used in the literature \citep{Russo2014-learning,Kandasamy2018-Parallelised,paria2020-flexible}.
    Then, we can extend $R_1$ as follows: 
    \begin{align}
        R_1 
        &= \sum_{t=1}^T \EE\left[ f(\*x^*) - V_t(\*x^*) \right] \\
        &\leq \sum_{t=1}^T \EE\left[ \bigl( f(\*x^*) - V_t(\*x^*) \bigr)_+ \right] && \bigl(\because f(\*x^*) - V_t(\*x^*) \leq \bigl( f(\*x^*) - V_t(\*x^*) \bigr)_+ \bigr) \\
        &\leq \sum_{t=1}^T \EE\left[ \sum_{\*x \in \cX} \bigl( f(\*x) - V_t(\*x) \bigr)_+ \right] && \left(\because \bigl( f(\*x^*) - V_t(\*x^*) \bigr)_+ \leq \sum_{\*x \in \cX} \bigl( f(\*x) - V_t(\*x) \bigr)_+ \right)\\
        &= \sum_{t=1}^T \sum_{\*x \in \cX} \EE_{\cD_{t-1}, \zeta_t} \left[ \EE \left[ \bigl( f(\*x) - V_t(\*x) \bigr)_+ \mid \cD_{t-1}, \zeta_t \right] \right]. 
    \end{align}
    Here, $\bigl( f(\*x) - V_t(\*x) \bigr) \mid \cD_{t-1}, \zeta_t$ follows $\cN(-\zeta_t^{1/2} \sigma_{t-1}(\*x), \sigma_{t-1}^2(\*x))$.
    Thus, by using Eq.~\eqref{eq:Gauss_plus_expectation}, we can obtain the following:
    \begin{align}
        R_1
        &\leq \sum_{t=1}^T \sum_{\*x \in \cX} \EE_{\cD_{t-1}, \zeta_t} \left[ \frac{\sigma_{t-1}(\*x)}{\sqrt{2 \pi}} \exp \left\{ - \frac{\zeta_t}{2} \right\} \right] && \bigl( \because \text{Eq.~\eqref{eq:Gauss_plus_expectation}} \bigr)\\
        &\leq \frac{|\cX|}{\sqrt{2 \pi}} \sum_{t=1}^T \EE_{\zeta_t} \left[ \exp \left\{ - \frac{\zeta_t}{2} \right\} \right] && \bigl( \because \sigma_{t-1}(\*x) \leq \sigma_0 (\*x) = 1 \bigr) \\
        &= \frac{|\cX|}{\sqrt{2 \pi}} \sum_{t=1}^T M_t(- 1/2). \label{eq:R_1_bound} && \bigl(\because \text{Definition of MGF} \bigr)
    \end{align}

    Second, we expand $R_2$ as follows:
    \begin{align}
        R_2 
        &= \sum_{t=1}^T \EE \left[ V_t(\*x_t) - f(\*x_t) \right] \\
        &= \sum_{t=1}^T \EE_{\cD_{t-1}, \zeta_t} \left[ \EE \left[ V_t(\*x_t) - f(\*x_t) \mid \cD_{t-1}, \zeta_t \right] \right] \\
        &= \sum_{t=1}^T \EE_{\cD_{t-1}, \zeta_t} \left[ V_t(\*x_t) - \mu_{t-1}(\*x_t) \right] \\
        &= \sum_{t=1}^T \EE_{\cD_{t-1}, \zeta_t} \left[ \zeta_t^{1/2} \sigma_{t-1}(\*x_t) \right] \\
        &= \EE \left[ \sum_{t=1}^T \zeta_t^{1/2} \sigma_{t-1}(\*x_t) \right] \\
        &\leq \EE \left[ \sqrt{ \sum_{t=1}^T \zeta_t \sum_{t=1}^T \sigma_{t-1}^2(\*x_t) } \right]. && \bigl( \because \text{Cauchy--Schwarz  inequality} \bigr)
    \end{align}
    Then, as with Lemma~5.4 in \citep{Srinivas2010-Gaussian}, $\sqrt{\sum_{t=1}^T \sigma_{t-1}^2(\*x_t) } \leq \sqrt{C_1 \gamma_T}$ with probability 1, where $C_1 = 2 / \log(1 + \sigma^{-2})$.
    Hence, 
    \begin{align}
        R_2 
        &\leq \EE \left[ \sqrt{ \sum_{t=1}^T \zeta_t } \right] \sqrt{C_1 \gamma_T} \\
        &\leq \sqrt{ \EE \left[ \sum_{t=1}^T \zeta_t \right] } \sqrt{C_1 \gamma_T} && \bigl( \because \text{Jensen's  inequality} \bigr) \\
        &= \sqrt{ \sum_{t=1}^T \EE \left[ \zeta_t \right] } \sqrt{C_1 \gamma_T}. \label{eq:R_2_bound} 
    \end{align}
    Consequently, combining Eqs.~\eqref{eq:R_1_bound} and \eqref{eq:R_2_bound}, we can obtain the theorem for (i).

    Then, we consider the case (ii).
    As with the proof of Theorem~\ref{theo:BCR_GPUCB}, purely for the sake of analysis, we used a set of discretization $\cX_t \subset \cX$ for $t \geq 1$.
    For any $t \geq 1$, let $\cX_t \subset \cX$ be a finite set with each dimension equally divided into $\tau_t = bdr t^2 \bigl(\sqrt{\log (ad)} + \sqrt{\pi} / 2 \bigr)$.
    Thus, $|\cX_t| = \tau_t^d$.
    In addition, we define $[\*x]_t$ as the nearest points in $\cX_t$ of $\*x \in \cX$.

    Then, we decompose BCR as follows:
    \begin{align}
        {\rm BCR}_T 
        &= \sum_{t=1}^T \EE \left[ f(\*x^*) - f([\*x^*]_t) + f([\*x^*]_t) - V_t([\*x^*]_t) + V_t([\*x^*]_t) - V_t(\*x_t) + V_t(\*x_t) - f(\*x_t) \right] \\
        &\leq \sum_{t=1}^T \EE \left[ f(\*x^*) - f([\*x^*]_t) + f([\*x^*]_t) - V_t([\*x^*]_t) + V_t(\*x_t) - f(\*x_t) \right],
    \end{align}
    where $\EE \bigl[ V_t([\*x^*]_t) - V_t(\*x_t) \bigr] \leq 0$ since $\*x_t = \argmax_{\*x \in \cX} V_t(\*x)$.
    Furthermore, we obtain the following:
    \begin{align*}
        \sum_{t=1}^T \EE \left[f(\*x^*) - f([\*x^*]_t) \right]
        &\leq \sum_{t=1}^T \EE \left[ \sup_{\*x \in \cX} f(\*x) - f([\*x]_t) \right] \\
        &\leq \sum_{t=1}^T \frac{1}{t^2} && \left( \because \text{Lemma}~\ref{lem:discretized_error} \right) \\
        &\leq \frac{\pi^2}{6} && \left(\because \sum_{t=1}^\infty \frac{1}{t^2} = \frac{\pi^2}{6} \right)
    \end{align*}
    The remaining terms $\sum_{t=1}^T \EE\bigl[ f([\*x^*]_t) - V_t([\*x^*]_t) + V_t(\*x_t) - f(\*x_t) \bigr]$ can be bounded as with the case (i).
    That is, 
    \begin{align}
        \sum_{t=1}^T \EE\bigl[ f([\*x^*]_t) - V_t([\*x^*]_t) + V_t(\*x_t) - f(\*x_t) \bigr]
        &\leq \frac{1}{\sqrt{2\pi}} \sum_{t=1}^T |\cX_t| M_t(-1/2) + \sqrt{C_1 \sum_{t=1}^T \EE[\zeta_t] \gamma_T}.
    \end{align}
    By substituting $|\cX_t| = (\tau_t)^d$, we conclude the proof.
\end{proof}

\subsection{Example of Distributions for Sub-linear BCR Bounds}
\label{app:examples_RGPUCB_dist}
Next, we show examples of the distributions that can achieve sub-linear regret bound using Theorem~\ref{theo:BCR_RGPUCB}.
Here, we discuss the case (ii) for infinite $\cX$.
The derivation for case (i) can be obtained easily by replacing $|\cX_t| = (\tau_t)^d$ with $|\cX|$.

Theorem~\ref{theo:BCR_RGPUCB} shows that BCR can be bounded by the expectation and MGF.
Roughly speaking, if we can control MGF as $M_t (- 1/2) = O(1 / (t^2 |\cX_t|) )$ and the expectation $\EE[\zeta_t] = O(\log t)$ by scheduling the parameter of the distribution of $\zeta_t$, we can obtain sub-linear regret bounds, which is similar to Theorem~\ref{theo:BCR_GPUCB}.
This is because we can bound the first term by $ \frac{1}{\sqrt{2 \pi}} \sum_{t=1}^T 1 / t^2 = \pi^{2} / (6 \sqrt{2 \pi})$ if the following inequality holds: 
\begin{align}
 |\cX_t| M_t(- 1 / 2) \leq \frac{1}{t^2}.
\end{align}
This inequality can be rewritten as,
\begin{align}
    M_t(- 1 / 2) \leq \frac{1}{|\cX_t| t^2}. \label{eq:condition_MGF}
\end{align}
Furthermore, the expectation  $\EE[\zeta_t] = O(\log t)$ is required to bound the second term.

The above requirements can be satisfied not only by the Gamma distribution proposed in \citep{berk2021-randomized} but also by many other distributions, such as two-parameter exponential and truncated normal distributions.
Their PDFs, MGFs, and expectations are listed in Table~\ref{tab:dist_properties}.
Note that although we describe only those three distributions, we can use other distributions, such as chi-squared distribution and truncated normal with other truncation, to obtain sub-linear regret bound from Theorem~\ref{theo:BCR_RGPUCB}.
Each scheduling of the parameter and resulting regret bound are shown as follows:

\begin{table}[!t]
    \centering
    \caption{PDFs, MGFs, and expectations for distributions of $\zeta$, where $\Gamma$ is the Gamma function, and $\phi$ and $\Phi$ are PDF and cumulative distribution function of standard normal distribution, respectively.}
    \begin{tabular}{c|c|c|c}
         & PDF & MGF: $M(-1/2)$ & Expectation  \\ \hline
         Gamma: $ (k, \theta)$ & $\frac{\zeta^{k-1} e^{- \zeta / \theta}}{\Gamma(k) \theta^k}$ & $(1 + \theta / 2)^{- k}$ & $k \theta$ \\ \hline
         Two-parameter exponential: $(s, \lambda)$ & $\lambda e^{- \lambda (\zeta - s)}$ for $\zeta \geq s \geq 0$ & $\frac{\lambda}{\lambda + 1 / 2} e^{- s / 2}$& $s + 1 / \lambda$ \\ \hline
         Truncated normal: $(m, s^2 = 1)$ &  \multirow{2}{*}{$\frac{\phi( (\zeta - m) / s)}{\Phi(1) - \Phi(-1)}$ for $m - s \leq \zeta \leq m + s$} & \multirow{2}{*}{$e^{- m / 2 + 1 / 8} \frac{\Phi(3/2) - \Phi(-1/2)}{\Phi(1) - \Phi(-1)} $} & \multirow{2}{*}{$m$} \\
         with truncation $[m - s, m + s]$ &&&
    \end{tabular}
    \label{tab:dist_properties}
\end{table}

 \paragraph{Gamma Distribution:} If we use $\zeta_t$ following Gamma distribution with $k_t$ and $\theta$, we require the following:
 \begin{align*}
     (1 + \theta / 2)^{-k_t} \leq \frac{1}{|\cX_t| t^2}.
 \end{align*}
 Therefore, we obtain the following:
 \begin{align*}
     k_t \geq \frac{\log ( |\cX_t| t^2  )}{\log(1 + \theta / 2)}.
 \end{align*}
 The above derivation is mostly the same as the proof of Theorem~1 in \citep{berk2021-randomized}.
 However, \citet{berk2021-randomized} bounded as $ \sqrt{\sum_{t=1}^T \EE[\zeta_t]} \leq \sqrt{T \EE[\max_{t \leq T} \zeta_t]}$.
 This operation is correct, but the evaluation of $\EE[\max_{t \leq T} \zeta_t]$ is slightly complicated.
 To bound this term, \citet{berk2021-randomized} used nearly equal, which is forbidden.
 On the other hand, we know the expectation $\EE[\zeta_t] = k_t \theta$ analytically.
 Let 
 \begin{align}
     k_t 
     &= \frac{\log ( |\cX_t| t^2  )}{\log(1 + \theta / 2)} \\
     &= \frac{ d\log \bigl( bdr t^2 \bigl(\sqrt{\log (ad)} + \sqrt{\pi} / 2 \bigr)  \bigr) + 2\log (t)}{\log(1 + \theta / 2)}.
 \end{align}
 Consequently, since $\EE[\zeta_t] = k_t \theta$ is monotone increasing, we obtain the following bound:
 \begin{align}
     {\rm BCR}_T 
     \leq \frac{\pi^2}{6} + \frac{\pi^2}{6\sqrt{2\pi}} + \sqrt{C_1 \gamma_t T \EE[\zeta_T]} 
     = \frac{\pi^2}{6} + \frac{\pi^2}{6\sqrt{2\pi}} + \sqrt{C_1 \gamma_t T \theta k_T}.
 \end{align}

\paragraph{Two-Parameter Exponential Distribution:} If we use $\zeta_t$ following two-parameter exponential distribution with $s_t$ and $\lambda$, we requires the following:
 \begin{align*}
     \frac{\lambda}{\lambda + 1 / 2} e^{- s_t / 2} \leq e^{- s_t / 2} \leq \frac{1}{|\cX_t| t^2},
 \end{align*}
 where we bound $\frac{\lambda}{\lambda + 1 / 2} < 1$ so that the condition for $\lambda$ to maintain $s_t \geq 0$ will be eliminated.
 Therefore, we obtain the following:
 \begin{align*}
     s_t \geq 2 \log \left( |\cX_t| t^2 \right).
 \end{align*}
 Hence, by setting $s_t = 2 \log \left( |\cX_t| t^2 \right) = 2d\log \bigl( bdr t^2 \bigl(\sqrt{\log (ad)} + \sqrt{\pi} / 2 \bigr)  \bigr) + 4\log (t)$, since $\EE[\zeta_t] = s_t + 1/\lambda$ is monotone increasing, we obtain the following bound:
 \begin{align}
     {\rm BCR}_T 
     \leq \frac{\pi^2}{6} + \frac{\pi^2}{6\sqrt{2\pi}} + \sqrt{C_1 \gamma_t T \EE[\zeta_T]} 
     = \frac{\pi^2}{6} + \frac{\pi^2}{6\sqrt{2\pi}} + \sqrt{C_1 \gamma_t T (s_T + 1/\lambda)}.
 \end{align}

\paragraph{Truncated Normal Distribution:} For brevity, we use a truncated normal distribution with $s^2 = 1$ and truncation $[m - s, m + s]$.
We schedule $m_t$ so that the following inequality holds:
\begin{align*}
    e^{- m / 2 + 1 / 8} \frac{\Phi(3/2) - \Phi(-1/2)}{\Phi(1) - \Phi(-1)} < e^{- m_t / 2 + 1 / 8} < \frac{1}{|\cX_t| t^2},
\end{align*}
where we bound $\frac{\Phi(1) - \Phi(0)}{\Phi(1/2) - \Phi(-1/2)} < 1$ for simplicity.
Then, we obtain the following condition:
\begin{align*}
    m_t \geq 2 \log \left( |\cX_t| t^2 \right) + 1/4.
\end{align*}
Furthermore, the lower bound $m_t - s > 0$ should be satisfied.
Hence, by setting $m_t = 2 \log \left( |\cX_t| t^2 \right) + 1 = 2d\log \bigl( bdr t^2 \bigl(\sqrt{\log (ad)} + \sqrt{\pi} / 2 \bigr)  \bigr) + 4\log (t) + 1$, since $\EE[\zeta_t] = m_t$ is monotone increasing, we obtain the following bound:
\begin{align}
     {\rm BCR}_T 
     \leq \frac{\pi^2}{6} + \frac{\pi^2}{6\sqrt{2\pi}} + \sqrt{C_1 \gamma_t T \EE[\zeta_T]} 
     = \frac{\pi^2}{6} + \frac{\pi^2}{6\sqrt{2\pi}} + \sqrt{C_1 \gamma_t T m_T}.
 \end{align}

\section{Proofs of Lemmas~\ref{lem:bound_srinivas} and \ref{lem:bound_RGPUCB}}

\subsection{Proof of Lemma~\ref{lem:bound_srinivas}}
\label{app:proof_srinivas}
The main difference between Lemma~\ref{lem:bound_srinivas} and \citep[Lemma 5.1 in ][]{Srinivas2010-Gaussian} is that Lemma~\ref{lem:bound_srinivas} does not consider the union bound for all $t \geq 1$. 
That is, \citep[Lemma 5.1 in ][]{Srinivas2010-Gaussian} bounds the following probability:
\begin{align*}
    \Pr \left( f(\*x) \leq \mu_{t-1}(\*x) + \beta^{1/2}_t \sigma_{t-1}(\*x), \forall \*x \in \cX, \forall t \geq 1 \right).
\end{align*}
On the other hand, our Lemma~\ref{lem:bound_srinivas} bounds the following probability with one fixed $t$:
\begin{align*}
    \myPr_t \left( f(\*x) \leq \mu_{t-1}(\*x) + \beta^{1/2}_{\delta} \sigma_{t-1}(\*x), \forall \*x \in \cX\right),
\end{align*}
where $\beta^{1/2}_{\delta}$ does not depend on $t$ as described below.

Assume the same condition as in Lemma~\ref{lem:bound_srinivas}, and thus, $\beta_\delta = 2\log\bigl(|\cX| / (2\delta)\bigr)$. 
Then, for all $\*x \in \cX$ and $\cD_{t-1}$, we see that
\begin{align*}
    \myPr_t \left( f(\*x) > \mu_{t-1}(\*x) + \beta^{1/2}_{\delta} \sigma_{t-1}(\*x) \right) \leq \frac{\delta}{|\cX|},
\end{align*}
where we use an elementary result of Gaussian distribution shown in Lemma~\ref{lem:Gauss_tail_bound} in Appendix~\ref{app:lemmas}. 
Then, we can obtain the following bound:
\begin{align*}
    \myPr_t \left( f(\*x) > \mu_{t-1}(\*x) + \beta^{1/2}_{\delta} \sigma_{t-1}(\*x), \exists \*x \in \cX \right)
    &\leq \sum_{\*x \in \cX} \myPr_t \left( f(\*x) > \mu_{t-1}(\*x) + \beta^{1/2}_{\delta} \sigma_{t-1}(\*x) \right) \\
    &\leq \delta.
\end{align*}
Therefore, 
\begin{align*}
    \myPr_t \left( f(\*x) \leq \mu_{t-1}(\*x) + \beta^{1/2}_{\delta} \sigma_{t-1}(\*x), \forall \*x \in \cX \right) = 1 - \myPr_t \left( f(\*x) > \mu_{t-1}(\*x) + \beta^{1/2}_{\delta} \sigma_{t-1}(\*x), \exists \*x \in \cX \right) \geq 1 - \delta,
\end{align*}
which concludes the proof.

\subsection{Detailed Proof of Lemma~\ref{lem:bound_RGPUCB}}
\label{app:proof_lemma_IRGPUCB}

This section provides detailed proof of the following Lemma~\ref{lem:bound_RGPUCB}:
\begin{replemma}{lem:bound_RGPUCB}
    Let $f \sim \cG \cP (0, k)$, where $k$ is a stationary kernel and $k(\*x, \*x) = 1$, and $\cX$ be finite.
    Assume that $\zeta$ follows a two-parameter exponential distribution with $s = 2 \log (|\cX| / 2)$ and $\lambda = 1/2$.
    Then, the following inequality holds:
    \begin{align*}
        \EE[f(\*x^*)] \leq \EE \left[\max_{\*x \in \cX} \mu_{t-1}(\*x) + \zeta^{1/2}_t \sigma_{t-1}(\*x) \right],
    \end{align*}
    for all $t \geq 1$.
\end{replemma}
\begin{proof}
    It suffices to show that, for an arbitrary $\cD_{t-1}$, 
    \begin{align*}
        \EE_t[f(\*x^*) | \cD_{t-1}] \leq \EE_t \left[\max_{\*x \in \cX} \mu_{t-1}(\*x) + \zeta^{1/2}_t \sigma_{t-1}(\*x) \right],
    \end{align*}
    since the claim of Lemma~\ref{lem:bound_RGPUCB} can be rewritten as follows:
    \begin{align*}
        &\EE_{\cD_{t-1}}[ \EE_t[f(\*x^*) ]] \\
        &\leq \EE_{\cD_{t-1}} \left[ \EE_t \left[ \max_{\*x \in \cX} \mu_{t-1}(\*x) + \zeta^{1/2} \sigma_{t-1}(\*x) \right] \right].
    \end{align*}
    Then, fix the dataset $\cD_{t-1}$ and $\delta \in (0, 1)$.
    From Lemma~\ref{lem:bound_srinivas}, we can see that
    \begin{align*}
        \myPr_t \left( f(\*x^*) \leq \max_{\*x \in \cX} \mu_{t-1}(\*x) + \beta^{1/2}_{\delta} \sigma_{t-1}(\*x) \right) \geq 1 - \delta,
    \end{align*}
    where $\beta_{\delta} =  2 \log (|\cX| / (2\delta))$.
    Note that the probability is taken only with $f(\*x^*)$ since $\cD_{t-1}$ is fixed.
    Using the cumulative distribution function $F_t(\cdot) \coloneqq \myPr_t(f(\*x^*) \leq \cdot)$ and its inverse function $F^{-1}_t$, we can rewrite
    \begin{align*}
        F_t \left( \max_{\*x \in \cX} \mu_{t-1}(\*x) + \beta^{1/2}_{\delta} \sigma_{t-1}(\*x) \right) &\geq 1 - \delta \\ 
        \Longleftrightarrow \max_{\*x \in \cX} \mu_{t-1}(\*x) + \beta^{1/2}_{\delta} \sigma_{t-1}(\*x) &\geq F^{-1}_t ( 1 - \delta ),
    \end{align*}
    since $F^{-1}_t$ is a monotone increasing function.
    Since $\delta \in (0, 1)$ is arbitrary, we can see that
    \begin{align*}
         \Pr \biggl( \max_{\*x \in \cX} \mu_{t-1}(\*x) + \beta^{1/2}_{U} \sigma_{t-1}(\*x) \geq F^{-1}_t ( 1 - U ) \biggr) = 1,
    \end{align*}
    where $U \sim {\rm Uni}(0, 1)$ and probability is taken by the randomness of $U$.
    Then, because of the basic property of the expectation $\EE[X] \leq \EE[Y]$ if $\Pr(X \leq Y) = 1$, we can obtain,
    \begin{align*}
         \EE_{U} \left[ \max_{\*x \in \cX} \mu_{t-1}(\*x) + \beta^{1/2}_{U} \sigma_{t-1}(\*x)   \right]
         &\geq \EE_{U} \bigl[ F^{-1}_t ( 1 - U )   \bigr] \\
         &= \EE_{U} \bigl[ F^{-1}_t ( U )   \bigr],
    \end{align*}
    where we use the fact that $1 - U$ also follows ${\rm Uni}(0, 1)$.
    Since $F^{-1}_t ( U )$ and $f(\*x^*)$ are identically distributed as with the inverse transform sampling, we obtain
    \begin{align*}
         \EE_{U} \left[ \max_{\*x \in \cX} \mu_{t-1}(\*x) + \beta^{1/2}_{U} \sigma_{t-1}(\*x)   \right]
         \geq \EE_t \bigl[ f(\*x^*) \bigr].
    \end{align*}
    Then, $\beta_U$ can be decomposed as follows:
    \begin{align*}
        \beta_U = 2 \log (|\cX| / 2) - 2 \log(U).
    \end{align*}
    From the inverse transform sampling, $ - 2 \log(U) \sim {\rm Exp} (\lambda = 1/2)$.
    Hence, $\beta_U$ follows a two-parameter exponential distribution with $s = 2 \log (|\cX| / 2)$ and $\lambda = 1/2$.
    Therefore, we obtain the following:
    \begin{align*}
         \EE_t \left[ \max_{\*x \in \cX} \mu_{t-1}(\*x) + \zeta^{1/2} \sigma_{t-1}(\*x) \right]
         \geq \EE_t \bigl[ f(\*x^*)  \bigr].
    \end{align*}
\end{proof}

\section{Proof of Theorem~\ref{theo:BCR_IRGPUCB_discrete}}
\label{app:proof_discrete}

Using Lemma~\ref{lem:bound_RGPUCB}, we can obtain the proof of the following theorem for the finite input domain:
\begin{reptheorem}{theo:BCR_IRGPUCB_discrete}
    Let $f \sim \cG \cP (0, k)$, where $k$ is a stationary kernel and $k(\*x, \*x) = 1$, and $\cX$ be finite.
    Assume that $\zeta_t$ follows a two-parameter exponential distribution with $s = 2 \log (|\cX| / 2)$ and $\lambda = 1/2$ for any $t \geq 1$.
    Then, by running IRGP-UCB with $\zeta_t$, BCR can be bounded as follows: 
    \begin{align*}
        {\rm BCR}_T \leq \sqrt{C_1 C_2 T \gamma_T},
    \end{align*}
    where $C_1 \coloneqq 2 / \log(1 + \sigma^{-2})$ and $C_2 \coloneqq 2 + s$. 
\end{reptheorem}
\begin{proof}
    From Lemma~\ref{lem:bound_RGPUCB}, we obtain
    \begin{align}
        {\rm BCR}_T
        &= \sum_{t=1}^T \EE[f(\*x^*) - f(\*x_t)] \\
        &= \sum_{t=1}^T \EE[f(\*x^*) - V_t(\*x_t) + V_t(\*x_t) - f(\*x_t)] \\
        &\leq \sum_{t=1}^T \EE[V_t(\*x_t) - f(\*x_t)].
    \end{align}
    Then, we can easily derive the bound as follows:
    \begin{align}
        {\rm BCR}_T
        &\leq \sum_{t=1}^T \EE[V_t(\*x_t) - f(\*x_t)] \\
        &= \sum_{t=1}^T \EE_{\cD_{t-1}, \zeta_t} [ \EE [V_t(\*x_t) - f(\*x_t) | \cD_{t-1}, \zeta_t]] \\
        &= \sum_{t=1}^T \EE_{\cD_{t-1}} [ \EE_{\zeta_t} [V_t(\*x_t) ] - \mu_{t-1}(\*x_t)] \\
        &= \sum_{t=1}^T \EE_{\cD_{t-1}, \zeta_t} [ \zeta_t^{1/2} \sigma_{t-1}(\*x_t) ] \\
        &= \EE \left[ \sum_{t=1}^T \zeta_t^{1/2} \sigma_{t-1}(\*x_t) \right] \\
        &\leq \EE \left[\sqrt{ \sum_{t=1}^T \zeta_t \sum_{t=1}^T \sigma_{t-1}^2 (\*x_t) } \right] && \bigl( \because \text{Cauchy--Schwarz  inequality} \bigr) \\
        &\leq \EE \left[\sqrt{ \sum_{t=1}^T \zeta_t } \right] \sqrt{C_1 \gamma_T} && \bigl( \because \text{Lemma~5.4 in \citep{Srinivas2010-Gaussian}} \bigr) \\
        &\leq \sqrt{ \sum_{t=1}^T \EE [ \zeta_t ]} \sqrt{C_1 \gamma_T} && \bigl( \because \text{Jensen's inequality} \bigr) \\
        &= \sqrt{C_1 C_2 T \gamma_T}, && \bigl( \because \text{Definition of $\zeta_t$} \bigr)
    \end{align}
    where $C_2 = \EE [ \zeta_t ] = 2 + 2\log(|\cX|/2)$.
    This concludes the proof.
\end{proof}

\begin{corollary}
    Assume the same condition as in Theorem~\ref{theo:BCR_IRGPUCB_discrete}.
    Fix a required accuracy $\eta > 0$.
    Then, ${\rm BSR}_T \leq \eta$ by at most $T$ function evaluations, where $T$ is the smallest positive integer satisfying the following inequality:
    \begin{align*}
        \sqrt{\frac{C_1 C_2 \gamma_T}{T}} \leq \eta,
    \end{align*}
    where $C_1 \coloneqq 2 / \log(1 + \sigma^{-2})$ and $C_2 \coloneqq 2 + 2 \log (|\cX| / 2)$.
\end{corollary}
\begin{proof}
    This BCR bound in Theorem~\ref{theo:BCR_IRGPUCB_discrete} immediately suggests that the following:
    \begin{align*}
        {\rm BSR}_T \leq \frac{{\rm BCR}_T }{T} \leq \sqrt{ \frac{C_1 C_2 \gamma_T}{T}}.
    \end{align*}
    Therefore, if $\sqrt{ C_1 C_2 \gamma_T / T} < \eta$, then ${\rm BSR}_T < \eta$.
\end{proof}

\section{Proof of Theorem~\ref{theo:BCR_IRGPUCB_continuous}}
\label{app:proof_continuous}

\begin{reptheorem}{theo:BCR_IRGPUCB_continuous}
    Let $f \sim \cG \cP (0, k)$, where $k$ is a stationary kernel, $k(\*x, \*x) = 1$, and Assumption~\ref{assump:continuous_X} holds.
    Assume that $\zeta_t$ follows a two-parameter exponential distribution with $s_t = 2d \log(bdr t^2 \bigl( \sqrt{\log (ad)} + \sqrt{\pi} / 2\bigr)) - 2 \log 2$ and $\lambda = 1/2$ for any $t \geq 1$.
    Then, by running IRGP-UCB, BCR can be bounded as follows: 
    \begin{align*}
        {\rm BCR}_T \leq \frac{\pi^2}{6} + \sqrt{C_1 T \gamma_T (2 + s_T)},
    \end{align*}
    where $C_1 \coloneqq 2 / \log(1 + \sigma^{-2})$. 
\end{reptheorem}
\begin{proof}
    Purely for the sake of analysis, we used a set of discretization $\cX_t \subset \cX$ for $t \geq 1$.
    For any $t \geq 1$, let $\cX_t \subset \cX$ be a finite set with each dimension equally divided into $\tau_t = bdr t^2 \bigl(\sqrt{\log (ad)} + \sqrt{\pi} / 2 \bigr)$.
    Thus, $|\cX_t| = \tau_t^d$.
    In addition, we define $[\*x]_t$ as the nearest points in $\cX_t$ of $\*x \in \cX$.

    Then, we decompose BCR as follows:
    \begin{align}
        {\rm BCR}_T
        &= \sum_{t=1}^T \EE \Biggl[ f(\*x^*) - f( [\*x^*]_t ) + f( [\*x^*]_t ) - \max_{\*x \in \cX_t} V_t(\*x) + \max_{\*x \in \cX_t} V_t(\*x) - V_t(\*x_t) + V_t(\*x_t) - f(\*x_t) \Biggl].
    \end{align}
    Obviously, $\EE\bigl[\max_{\*x \in \cX_t} V_t(\*x) - V_t(\*x_t) \bigr] \leq 0$ from the definition of $\*x_t$.
    Furthermore, we observe the following:
    \begin{align*}
        \EE \left[ f( [\*x^*]_t ) - \max_{\*x \in \cX_t} V_t(\*x) \right] \leq \EE \left[ \max_{\*x \in \cX_t} f( \*x ) - \max_{\*x \in \cX_t} V_t(\*x) \right],
    \end{align*}
    which can be bounded above by $0$ by setting $s_t = 2\log(|\cX_t| / 2)$ and $\lambda = 1 / 2$ from Lemma~\ref{lem:bound_RGPUCB}.
    Therefore, we obtain the following:
    \begin{align}
        {\rm BCR}_T
        &\leq \sum_{t=1}^T \EE \left[ f(\*x^*) - f( [\*x^*]_t ) + V_t(\*x_t) - f(\*x_t) \right] \\
        &= \sum_{t=1}^T \EE \left[ f(\*x^*) - f( [\*x^*]_t ) \right] + \sum_{t=1}^T \EE \left[ V_t(\*x_t) - f(\*x_t) \right]. \label{eq:BCR_continuous_two_term_bound}
    \end{align}

    First, we consider the first term $ \sum_{t=1}^T \EE \bigl[ f(\*x^*) - f( [\*x^*]_t ) \bigr]$.
    From Lemma~\ref{lem:discretized_error}, we can obtain the following:
    \begin{align}
        \sum_{t=1}^T \EE \left[ f(\*x^*) - f( [\*x^*]_t ) \right] 
        &\leq \sum_{t=1}^T \EE \left[ \sup_{\*x \in \cX} f(\*x) - f( [\*x]_t ) \right] \\
        &\leq \sum_{t=1}^T \frac{1}{t^2} && \bigl( \because \text{ Lemma~\ref{lem:discretized_error} } \bigr) \\
        &\leq \frac{\pi^2}{6}, && \left(\because \sum_{t=1}^\infty \frac{1}{t^2} = \frac{\pi^2}{6} \right) \label{eq:discrete_points_distance_bound}
    \end{align}
    where $u_t$ in Lemma~\ref{lem:discretized_error} corresponds to $t^2$.

    The second term is bounded as follows:
    \begin{align}
        \sum_{t=1}^T \EE \left[ V_t(\*x_t) - f(\*x_t) \right]
        &\leq \sqrt{ \sum_{t=1}^T \EE[\zeta_t] C_1 \gamma_T }, && \bigl(\because \text{See the proof of Theorem~\ref{theo:BCR_IRGPUCB_discrete}} \bigr) \label{eq:continouos_bound_width_bound}
    \end{align}
    where 
    \begin{align}
        \EE[\zeta_t]
        &= 2 + 2\log(|\cX_t| / 2) \\
        &= 2 + 2d \log(bdr t^2 \bigl( \sqrt{\log (ad)} + \sqrt{\pi} / 2\bigr)) - 2 \log 2, 
    \end{align}
    which is monotone increasing.
    Therefore, we obtain the following:
    \begin{align}
        \sum_{t=1}^T \EE \left[ V_t(\*x_t) - f(\*x_t) \right]
        &\leq \sqrt{ C_1 T \gamma_T \EE[\zeta_T] },
    \end{align}
    where $\EE[\zeta_T] = 2 + s_T = 2 + 2d \log(bdr T^2 \bigl( \sqrt{\log (ad)} + \sqrt{\pi} / 2\bigr)) - 2 \log 2$.
    Consequently, combining Eqs.~\eqref{eq:BCR_continuous_two_term_bound}, \eqref{eq:discrete_points_distance_bound}, and \eqref{eq:continouos_bound_width_bound} concludes the proof.
\end{proof}

\begin{corollary}
    Assume the same condition as in Theorem~\ref{theo:BCR_IRGPUCB_continuous}.
    Fix a required accuracy $\eta > 0$.
    Then, by running IRGP-UCB with $s_\eta = 2d \log(2bdr \bigl( \sqrt{\log (ad)} + \sqrt{\pi} / 2\bigr) / \eta) - 2 \log 2$ and $\lambda = 1 / 2$, ${\rm BSR}_T \leq \eta$ by at most $T$ function evaluations, where $T$ is the smallest positive integer satisfying the following inequality:
    \begin{align*}
        \sqrt{\frac{C_1 (2 + s_\eta) \gamma_T}{T}} \leq \frac{\eta}{2},
    \end{align*}
    where  $C_1 \coloneqq 2 / \log(1 + \sigma^{-2})$.
\end{corollary}
\begin{proof}
    We obtain the upper bound of BSR via the upper bound of BCR as follows:
    \begin{align}
        {\rm BSR}_T \leq \frac{{\rm BCR}_T}{T}.
    \end{align}
    Let $\tau_t = 2bdr \bigl( \sqrt{\log (ad)} + \sqrt{\pi} / 2\bigr) / \eta$.
    Until the derivation of Eq.~\eqref{eq:BCR_continuous_two_term_bound}, the same derivation as in Theorem~\ref{theo:BCR_IRGPUCB_continuous} can be applied.
    That is, 
    \begin{align}
        {\rm BCR}_T
        &\leq \sum_{t=1}^T \EE \left[ f(\*x^*) - f( [\*x^*]_t ) \right] + \sum_{t=1}^T \EE \left[ V_t(\*x_t) - f(\*x_t) \right]. 
    \end{align}
    Then, for the first term, we can obtain the following
    \begin{align}
        \sum_{t=1}^T \EE \left[ f(\*x^*) - f( [\*x^*]_t ) \right]
        &\leq \sum_{t=1}^T \frac{\eta}{2} && \bigl( \because \text{ Lemma~\ref{lem:discretized_error} } \bigr) \\
        &\leq \frac{\eta}{2} T.
    \end{align}
    For the second term, 
    \begin{align}
        \sum_{t=1}^T \EE \left[ V_t(\*x_t) - f(\*x_t) \right]
        &\leq \sqrt{ C_1 \sum_{t=1}^T \EE[\zeta_t] \gamma_T },
    \end{align}
    where, from the assumption, $\EE [\zeta_t] = 2 + s_\eta$ for any $t \geq 1$.
    Thus, we can derive the bound of BSR:
    \begin{align}
        {\rm BSR}_T \leq \frac{\eta}{2} + \sqrt{\frac{C_1 (2 + s_\eta) \gamma_T}{T}}.
    \end{align}
    Hence, we obtain the smallest integer $T$ that satisfies ${\rm BSR}_T \leq \eta$ by arranging the following inequality:
    \begin{align*}
        \frac{\eta}{2} + \sqrt{\frac{C_1 (2 + s_\eta) \gamma_T}{T}} \leq \eta.
    \end{align*}
\end{proof}

\section{Auxiliary Lemmas}
\label{app:lemmas}

For convenience, we here show the assumption again:
\begin{repassumption}{assump:continuous_X}
    Let $\cX \subset [0, r]^d$ be a compact and convex set, where $r > 0$.
    Assume that the kernel $k$ satisfies the following condition on the derivatives of sample path $f$.
    There exists the constants $a, b > 0$ such that,
    \begin{align*}
        \Pr \left( \sup_{\*x \in \cX} \left| \frac{\partial f}{\partial \*x_j} \right| > L \right) \leq a \exp \left( - \left(\frac{L}{b}\right)^2 \right),\text{ for } j \in [d],
    \end{align*}
    where $[d] = \{1, \dots, d\}$.
\end{repassumption}

Then, from Assumption~\ref{assump:continuous_X}, we obtain several lemmas.
First, we show the upper bound of the supremum of the partial derivatives.
This result is tighter than Lemma~12 in \citep{Kandasamy2018-Parallelised}.
Note that, in the proof of Lemma~12 in \citep{Kandasamy2018-Parallelised}, there is a typo that $\Pr (L \geq t)$ is bounded above by $a \exp\{ t^2 / b^2 \}$, which should be $a d \exp\{ - t^2 / b^2 \}$.
\begin{lemma}
    Let $f \sim \cG \cP (0, k)$ and Assumption~\ref{assump:continuous_X} holds.
    Let the supremum of the partial derivatives $L_{\rm max} \coloneqq \sup_{j \in [d]} \sup_{\*x \in \cX} \left| \frac{\partial f}{\partial x_j} \right|$.
    Then, $\EE[L_{\rm max}]$ can be bounded above as follows:
    \begin{align}
        \EE[L_{\rm max}] \leq b \bigl( \sqrt{\log (ad)} + \sqrt{\pi} / 2 \bigr).
    \end{align}
    \label{lem:L_max_bound}
\end{lemma}
\begin{proof}
    From the assumption, using union bound, we obtain the bound of the probability,
    \begin{align}
        \Pr\left( L_{\rm max} > L \right) \leq \sum_{j=1}^d a \exp \left( - \left(\frac{L}{b}\right)^2 \right) = a d \exp \left( - \left(\frac{L}{b}\right)^2 \right). \label{eq:L_max_prob}
    \end{align}
    Then, using expectation integral identity, we can bound $\EE[L_{\rm max}]$ as follows:
    \begin{align}
        \EE[L_{\rm max}]
        &= \int_0^{\infty} \Pr\left( L_{\rm max} > L \right) {\rm d} L \\
        &\leq \int_0^{\infty} \min \{ 1, ad e^{- (L / b)^2} \} {\rm d} L && \bigl(\because \text{Eq.~\eqref{eq:L_max_prob}} \bigr) \\
        &= b \sqrt{\log (ad)} + \int_{b\sqrt{\log (ad)}}^{\infty} ad e^{- (L / b)^2} {\rm d} L \\
        &= b \sqrt{\log (ad)} + a b d \sqrt{\pi} \int_{b\sqrt{\log (ad)}}^{\infty} \frac{1}{\sqrt{2\pi (b^2/2)}} e^{- (L / b)^2} {\rm d} L \\
        &= b \sqrt{\log (ad)} + a b d \sqrt{\pi} \left( 1 - \Phi \left( \frac{b\sqrt{\log (ad)}}{ b / \sqrt{2}} \right) \right) \\
        &\leq b \sqrt{\log (ad)} + \frac{ b \sqrt{\pi}}{2}, && \bigl(\because \text{Lemma~\ref{lem:Gauss_tail_bound}} \bigr)
    \end{align}
    where $\Phi$ is a cumulative distribution function of the standard normal distribution.
\end{proof}

Next, we show the bound for the difference of discretized outputs:
\begin{lemma}
    Let $f \sim \cG \cP (0, k)$ and Assumption~\ref{assump:continuous_X} holds.
    Let $\cX_t \subset \cX$ be a finite set with each dimension equally divided into $\tau_t = bdr u_t \bigl( \sqrt{\log (ad)} + \sqrt{\pi} / 2 \bigr)$ for any $t \geq 1$.
    Then, we can bound the expectation of differences,
    \begin{align}
        \sum_{t=1}^T \EE \left[ \sup_{\*x \in \cX} f(\*x) - f( [\*x]_t ) \right] \leq \sum_{t=1}^T \frac{1}{u_t},
    \end{align}
    where $[\*x]_t$ is the nearest point in $\cX_t$ of $\*x \in \cX$.
    \label{lem:discretized_error}
\end{lemma}
\begin{proof}
    From the construction of $\cX_t$, we can obtain the upper bound of L1 distance between $\*x$ and $[\*x]_t$ as follows:
    \begin{align}
        \sup_{\*x \in \cX}\| \*x - [\*x]_t \|_1
        &\leq \frac{dr }{bdr u_t \bigl( \sqrt{\log (ad)} + \sqrt{\pi} / 2\bigr) } \\
        &= \frac{1}{b u_t \bigl( \sqrt{\log (ad)} + \sqrt{\pi} / 2\bigr)}.
        \label{eq:sup_L1dist_discretization}
    \end{align}
    Note that this discretization does not depend on any randomness and is fixed beforehand.

    Then, we obtain the following:
    \begin{align}
        \sum_{t=1}^T \EE \left[ \sup_{\*x \in \cX} f(\*x) - f( [\*x]_t ) \right] 
        &\leq \sum_{t=1}^T \EE \left[ L_{\rm max} \sup_{\*x \in \cX} \| \*x - [\*x]_t \|_1 \right] \\
        &\leq \sum_{t=1}^T \EE \left[ L_{\rm max} \right] \frac{1}{b u_t \bigl( \sqrt{\log (ad)} + \sqrt{\pi} / 2\bigr)} && \bigl(\because \text{Eq.~\eqref{eq:sup_L1dist_discretization}} \bigr) \\
        &\leq \sum_{t=1}^T \frac{1}{u_t}. && \bigl(\because \text{Lemma~\ref{lem:L_max_bound}} \bigr)
    \end{align}
\end{proof}


We used the following useful lemma:
\begin{lemma}[in Lemma 5.2 of \citep{Srinivas2010-Gaussian}]
    For $c > 0$, the survival function of the standard normal distribution can be bounded above as follows:
    \begin{align}
        1 - \Phi(c) \leq \frac{1}{2} \exp (- c^2 / 2).
    \end{align}
    \label{lem:Gauss_tail_bound}
\end{lemma}
\begin{proof}
    Let $R \sim \cN(0, 1)$.
    Then, we obtain the following:
    \begin{align}
        \Pr(R > c) &= \int_{c}^\infty \frac{1}{\sqrt{2 \pi}} \exp ( - r^2 / 2) {\rm d}r \\
        &= \exp (- c^2 / 2) \int_{c}^\infty \frac{1}{\sqrt{2 \pi}} \exp ( - r^2 / 2 + c^2 / 2) {\rm d}r \\
        &= \exp (- c^2 / 2) \int_{c}^\infty \frac{1}{\sqrt{2 \pi}} \exp ( - (r - c)^2 / 2 - rc + c^2) {\rm d}r \\
        &= \exp (- c^2 / 2) \int_{c}^\infty \frac{1}{\sqrt{2 \pi}} \exp ( - (r - c)^2 / 2 - c(r - c)) {\rm d}r \\
        &\leq \exp (- c^2 / 2) \int_{c}^\infty \frac{1}{\sqrt{2 \pi}} \exp ( - (r - c)^2 / 2) {\rm d}r && \left( \because \text{$e^{-c(r-c)} \leq 1$ since $r \geq c > 0$} \right) \\
        &= \frac{1}{2} \exp (- c^2 / 2).
    \end{align}
    %
\end{proof}

\end{document}